\documentclass[lettersize,journal]{IEEEtran}
\usepackage{amsmath,amsfonts}
\usepackage{algorithmic}
\usepackage{algorithm}
\usepackage{array}
\usepackage[caption=false,font=normalsize,labelfont=sf,textfont=sf]{subfig}
\usepackage{textcomp}
\usepackage{stfloats}
\usepackage{url}
\usepackage{verbatim}
\usepackage{graphicx}
\usepackage{cite}
\usepackage[table]{xcolor}
\usepackage[pagebackref=true,breaklinks=true,citecolor=blue,linkcolor=blue,colorlinks,urlcolor=blue,bookmarks=false]{hyperref}

\begin{document}

\title{Primitive-Driven Compositional Forensic Visual Prompting for Open-World Face Anti-Spoofing}

\author{Fangling~Jiang, Qi~Li,  Bing Liu,  Weining Wang, Quilin Huang, Zhenan~Sun, Ming-Hsuan Yang
\thanks{Fangling Jiang and Qi Li contributed equally to this study. Corresponding author: Qi~Li.}
\thanks{Fangling Jiang, Bing Liu and Quilin Huang are with The First Affiliated Hospital, the School of Computer Science, University of South China, Hengyang, 421001, China (email: \{jfl, bingliu, hql\}@usc.edu.cn)}
\thanks{Qi Li, Weining Wang and Zhenan Sun are with the New Laboratory of Pattern Recognition, MAIS, CASIA, Beijing, 100190, China (email: \{qli, weining.wang, znsun\}@nlpr.ia.ac.cn).}
\thanks{Ming-Hsuan Yang is with the Department of Computer Science and Engineering, University of California, Merced, CA, 95340, USA (email:mhyang@ucmerced.edu).
}

}



\maketitle

\begin{abstract}
	Open-world face anti-spoofing must address both covariate and semantic shifts: source and target domains differ in imaging conditions, while target domains contain diverse attack types absent from training. 
	Existing prompt-based approaches often express spoofing through category semantics or language guidance, which is effective for modeling high-level concepts but is less suited to explicitly capturing the evolving fine-grained and spatially heterogeneous forensic evidence of unseen attacks.
	Motivated by the hypothesis that many unseen attacks can be characterized by new combinations of recurring visual cues, we propose a compositional forensic visual prompt learning framework that operates entirely in the visual feature space. 
    Built on a frozen ViT-based vision foundation model, the framework employs patch-aware attention to refine a shared set of learnable micro-forensic primitives into localized forensic evidence units derived from image patches. Class-specific global contextual prompts then provide input-dependent routing weights that adaptively select and compose these primitives into compositional forensic visual prompts for real/spoof discrimination. The primitives are not assigned predefined semantic meanings; instead, their specialization and reuse emerge from shared parameterization and joint optimization across categories.
    Extensive experiments on nine open-world protocols demonstrate state-of-the-art performance, strong cross-domain generalization, and robust adaptation to unseen attacks. Further analyses in the frequency, spatial, feature, and primitive domains show that the learned primitives capture transferable and composable local forensic evidence, while global-context-guided routing produces input-dependent and discriminative primitive compositions. These results support visual primitive composition as an effective paradigm for detecting heterogeneous attacks in open-world face anti-spoofing.
\end{abstract}


\section{Introduction}
\begin{figure*}[tb]
	\centering
	\includegraphics[width=1\textwidth]{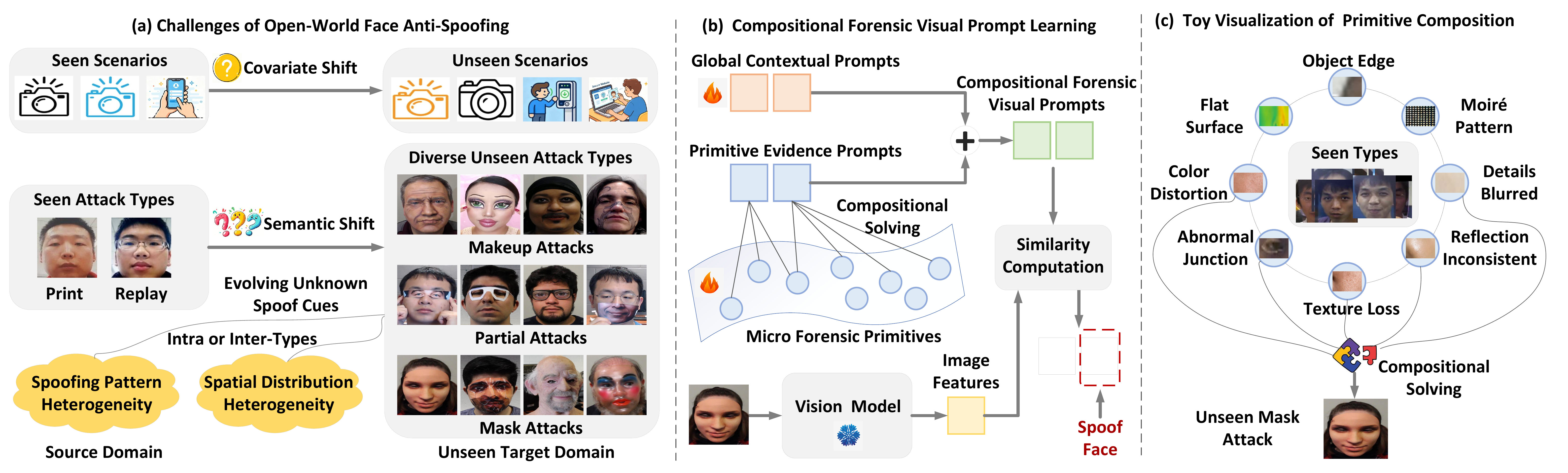}
	\caption{
		(a) Open-world face anti-spoofing exhibits severe covariate and semantic shifts, resulting in continuously evolving heterogeneous spoof cues in both pattern and spatial distribution.
		(b) Our framework learns reusable micro-forensic primitives and dynamically composes them using image-level physical priors for unified detection of heterogeneous attacks.
		(c) Motivating hypothesis: heterogeneous unseen attacks may be represented by new combinations of recurring visual forensic cues.
	}
	\label{moti}
\end{figure*}

\IEEEPARstart{F}{ace} anti-spoofing distinguishes real faces from presentation attacks and is essential to securing face recognition systems. Existing studies\cite{yu2021deep} have primarily investigated cross-scenario generalization, typically assuming that source and target domains share the same attack types (e.g., print and replay attacks) but differ in acquisition conditions, such as cameras, illumination, and backgrounds. Under this assumption, the dominant challenge is covariate shift, which is commonly addressed through domain adaptation\cite{jia2021unified,yang2026fine,li2025optimal,mao2025weighted} and domain generalization\cite{jia2020single,zhou2023instance,jung2025group,jiang2023adversarial}. Real-world face anti-spoofing, however, is inherently open-world\cite{jiang2024open,huang2024one,narayan2024hyp}. The asymmetric adversarial dynamics between attackers and defenders continually produce unseen attacks with distinct materials, geometric structures, and visual appearances, as illustrated in \figurename~\ref{moti}. 
Open-world continually evolving spoofing attacks exhibit substantial heterogeneity in both visual patterns and spatial distributions. Their forensic evidence spans multiple scales, ranging from global physical anomalies, such as three-dimensional geometric distortions, to local fine-grained artifacts. Such evidence varies markedly both across and within attack types, and may occur over the entire face or be confined to different facial regions. 
These attacks introduce semantic shifts beyond conventional covariate shift, causing models trained on seen attack types to generalize poorly\cite{jiang2024open,huang2024one,narayan2024hyp}. 

Recent studies exploit the transferable knowledge of large-scale vision-language models through textual or multimodal prompt learning\cite{srivatsan2023flip,yu2025multi,wang2024tf,hu2024fine,guo2025domain,zhang2026long,zhang2026cpg,liu2026icpe,liu2026dgpdl,liu2024bottom,guo2024style,cai2025me,ma2026clip}, achieving promising performance under both covariate and semantic shifts.
However, they rely on language semantics as an intermediary for visual knowledge representation, making them inherently limited in modeling heterogeneous unseen attacks. Text encoders emphasize category-level abstraction and may suppress low-level and high-frequency forensic evidence that is critical for face anti-spoofing during cross-modal alignment. Moreover, many spoof-specific cues, such as irregular specular reflections, material texture anomalies, and subtle frequency distortions, are difficult to describe precisely in natural language, while textual prompts typically provide global semantic supervision with limited spatial sensitivity. 
More importantly, fixed textual prompts used at inference time are difficult to adapt to continually emerging unseen attacks with diverse physical properties and forensic manifestations.

Motivated by this gap, we investigate a compositional visual diagnosis paradigm. 
Our working hypothesis is that many unseen attacks can be characterized by previously observed forensic cues appearing in new combinations rather than by an entirely new physical mechanism. For example, a silicone-mask attack can contain a combination of geometric rigidity, unusual reflection, and altered skin texture, while related cues may also occur individually in other seen presentation attacks, such as print and replay attacks. We therefore represent spoofing evidence with a shared set of learnable visual primitives and combine them adaptively for each input. We use the term \emph{compositional} in an operational sense: the class-specific representation is explicitly constructed as an input-dependent weighted combination of refined evidence primitives; the primitives are not assumed to correspond to manually defined semantic factors.

Accordingly, we propose a purely visual prompt learning framework based on micro-forensic primitive composition, as illustrated in \figurename~\ref{moti}. Unlike language-mediated prompting, the proposed framework learns prompts directly in the continuous visual feature space, enabling joint modeling of low-frequency global structures and high-frequency local artifacts that are difficult to describe linguistically. We initialize micro-forensic primitives as learnable vectors and refine them through patch-aware attention, allowing each primitive to specialize in a distinct type of reusable fine-grained evidence, such as geometric rigidity, abnormal reflections, and missing skin texture. In this way, heterogeneous spoofing traces are decomposed into elementary evidence units and organized within a shared primitive library, providing a structured basis for compositional diagnosis.
We further introduce a dynamic routing mechanism in which global contextual prompts encode image-level physical priors and serve as conditioning signals for primitive composition. Guided by the holistic physical characteristics of each input, the framework adaptively selects and combines task-relevant primitives to generate input-specific compositional forensic visual prompts for real/spoof discrimination.

The proposed framework avoids reliance on predefined attack categories or textual descriptions and establishes a unified reasoning paradigm for both seen and unseen attacks directly in the visual feature space. By learning compact real representations and organizing heterogeneous spoofing patterns within a shared primitive space, the framework provides an explicit intermediate representation in which local forensic evidence can be reused across attack types, thereby enabling compositional reasoning over unseen heterogeneous attacks and improving zero-shot generalization in open-world scenarios.

The main contributions of this work are:

\begin{itemize}

\item We formulate open-world face anti-spoofing from a compositional visual evidence perspective, representing heterogeneous attacks with input-dependent combinations of shared forensic evidence units rather than fixed attack-category descriptions.

\item We propose a purely visual prompt learning framework that organizes spoofing artifacts into reusable micro-forensic primitives and dynamically routes and composes them into input-adaptive compositional forensic visual prompts for unified inference across diverse seen and unseen attacks.

\item Extensive experiments on nine open-world protocols demonstrate that the proposed method achieves state-of-the-art performance under concurrent covariate and semantic shifts, together exhibiting strong zero-shot generalization to heterogeneous unseen attacks.
\end{itemize}

\section{Related Work}
\subsection{Cross-Domain Face Anti-Spoofing}
Cross-domain face anti-spoofing aims to improve generalization ability under unseen acquisition devices, illumination conditions, background environments, and attack types. Existing studies mainly follow four directions: domain adaptation, domain generalization, multimodal generalization, and open-set/one-class generalization. 

Domain adaptation methods exploit target domain data under unsupervised, semi-supervised, source-free, or test-time settings to reduce source and target discrepancies. Representative techniques include adversarial alignment\cite{jia2021unified,wang2020unsupervised}, pseudo-label learning\cite{yang2026fine,huang2023test}, optimal transport\cite{li2025optimal,mao2025weighted}, and self-supervised exploration\cite{liu2021casia}. In contrast, domain generalization methods learn transferable representations without accessing target domain data. Existing approaches employ domain-adversarial learning\cite{jia2020single,jiang2023adversarial,zhou2023instance}, knowledge distillation\cite{kong2024dual}, gradient or consistency regularization\cite{liu2025dual,le2024gradient}, meta-learning\cite{jia2021dual}, continual learning\cite{cai2025rehearsal}, and liveness-irrelevant feature disentanglement\cite{jung2025group,yang2024generalized,ma2024dual,liu2020disentangling}. Other studies improve generalization through data-centric or frequency-domain strategies, including physics-driven synthesis\cite{cai2024towards}, diffusion-based generation\cite{ge2024difffas}, and frequency-shortcut suppression\cite{cao2025towards}. Multimodal methods further exploit cross-modal alignment\cite{ma2026purify,lin2025reliable} and fine-grained textual guidance\cite{chen2025crossgap,li2025fine} to enhance cross-scenario robustness.
Beyond covariate shift, open-set\cite{jiang2024open,dong2021open} and one-class\cite{huang2024one,narayan2024hyp} methods explicitly address attacks absent from training. These approaches model real face distributions\cite{narayan2024hyp}, synthesize or extract spoofing cues\cite{jiang2024open,huang2024one}, or learn unknown spoof prompts\cite{jiang2026learning}. More recently, multimodal large language models have been introduced to improve both interpretability and generalization\cite{zhang2026intuition,zhang2026harnessing,zhang2025interpretable,ma2026pa,wang2025faceshield}.

Overall, existing methods have made substantial progress in mitigating cross-domain covariate shift. However, most of them still rely on domain-invariant representations, seen attack distributions, or language-semantic guidance, leaving the continually evolving heterogeneous attacks and their fine-grained forensic evidence insufficiently modeled in open-world scenarios.

\subsection{Prompt Learning in Face Anti-Spoofing}
Prompt learning-based face anti-spoofing methods exploit prior knowledge encoded in vision-language models\cite{Zhou_2022} to improve generalization under cross-domain and unseen-attack scenarios. Existing approaches can be broadly divided into textual and multimodal prompting.

Text prompt methods are pioneered by FLIP\cite{srivatsan2023flip}, which injects face anti-spoofing-related semantic information into visual representation learning through language guidance, thereby enhancing the generalization capability of real/spoof discriminative features. Building on this paradigm, subsequent studies have further designed fine-grained text prompts\cite{yu2025multi,wang2024tf,hu2024fine,guo2025domain,zhang2026long,zhang2026cpg,11194267}, instance and category prompts\cite{liu2026icpe}, domain prompts\cite{liu2026dgpdl,liu2024bottom}, and style-conditioned prompts\cite{guo2024style} to improve the descriptive capacity of textual prompts from the perspectives of attack attributes, domain discrepancies, style variations, and local semantics. 
Multimodal prompt\cite{cai2025me} methods further integrate visual, textual, and cross-modal alignment information\cite{ma2026clip}, improving the consistency and generalizability of vision-language representations. In addition, several works exploit text-to-image generation to expand the training distribution\cite{ko2025text}, or unknown spoof prompt learning\cite{huang2025slip} to improve generalization to unseen attacks. 

Overall, prompt learning-based methods demonstrate the value of foundation-model knowledge for generalizable face anti-spoofing. Nevertheless, most approaches still rely on textual descriptions, category semantics, or cross-modal alignment to represent spoofing patterns. Our approach is complementary: it discards the text encoder and treats learnable visual queries, refined directly against image patches, as candidate local evidence units. These evidence units are then combined by class-conditioned routing rather than being assigned fixed semantic labels. The distinction is therefore not simply text versus visual tokens; the goal is to expose an input-adaptive intermediate evidence space that can reuse local visual patterns across heterogeneous attack types.

\section{Proposed Method}
\begin{figure*}[tb]
	\centering
	\includegraphics[width=1\textwidth]{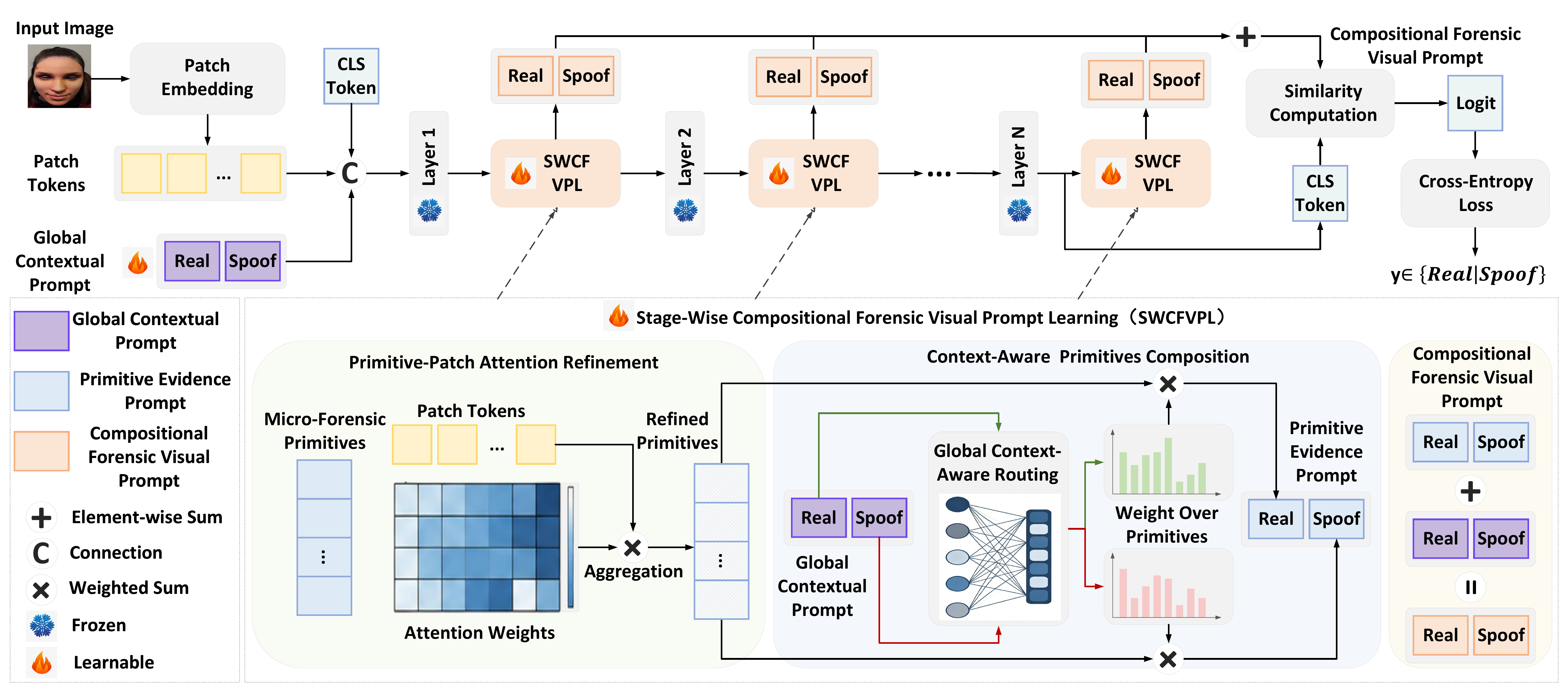}
	\caption{
		Overview of the proposed compositional forensic diagnosis framework for open-world face anti-spoofing. Built on a frozen ViT-based vision foundation model, the framework learns hierarchical purely visual prompts without language encoding. At each stage, micro-forensic primitives are first refined through patch-aware interactions to summarize reusable and composable fine-grained forensic units. These forensic units are then dynamically composed into class-specific primitive evidence prompts according to global priors and augmented with global contextual prompts to generate stage-wise compositional forensic visual prompts. Multi-stage prompts are finally aggregated for real/spoof discrimination. By representing heterogeneous attacks as dynamic compositions of reusable forensic evidence, the framework enables unified reasoning over diverse unseen attack types.
	}
	\label{figfr}
\end{figure*}
\subsection{Problem Formulation and Overall Architecture}
Open-world face anti-spoofing aims to learn a model that generalizes under concurrent covariate and semantic shifts. We consider a general setting with a single source domain
$
\mathcal{D}^{s}
=
\left\{
(\mathbf{x}_{i}^{s},\mathbf{y}_{i}^{s})
\right\}_{i=1}^{N^{s}}
$
for training and multiple unseen target domains
\(\{\mathcal{D}_{m}^{u}\}_{m=1}^{M}\) for evaluation, where \(\mathbf{x}_{i}^{s}\) and \(\mathbf{y}_{i}^{s}\) denote the \(i\)-th training sample and its label, respectively. Let \(\mathcal{Z}^{s}\) and \(\mathcal{Z}_{m}^{u}\) denote the acquisition-condition sets of the source domain and the \(m\)-th target domain, and let \(\mathcal{A}^{s}\) and \(\mathcal{A}_{m}^{u}\) denote their corresponding attack-type sets. The open-world setting is characterized by
\begin{equation}\label{eqopen_world_shift}
	\mathcal{Z}_{m}^{u}\neq\mathcal{Z}^{s},
	\qquad
	\mathcal{A}_{m}^{u}\setminus\mathcal{A}^{s}\neq\emptyset,
	\qquad
	m\in\{1,\ldots,M\},
\end{equation}
where the first condition represents acquisition-induced covariate shift, while the second indicates that the target domain contains attack types absent during training.

Unlike methods that represent class concepts through textual prompts, we learn purely visual prompts directly in the visual feature space to capture forensic evidence across multiple granularities, from image-level physical context to localized artifacts. We formulate open-world face anti-spoofing as a compositional forensic diagnosis problem, in which discriminative evidence is organized into a shared library of learnable visual primitives and dynamically composed into input-specific prompts. This formulation promotes cross-category reuse of local forensic evidence and enables unified detection of both seen and unseen attacks.

As illustrated in \figurename~\ref{figfr}, the proposed framework performs visual prompt learning on a frozen ViT-based vision foundation model. We simply retain the visual encoder of pretrained CLIP as the vision foundation model, discard its text encoder, and freeze all backbone parameters during training. Since shallow layers primarily capture local textures, boundaries, and high-frequency artifacts, whereas deeper layers encode structural, geometric, and higher-level physical discrepancies, we partition the ViT backbone into $L$ hierarchical stages and inject task-specific forensic knowledge through stage-wise compositional forensic visual prompts.

At each stage $l$, a compositional forensic visual prompt $\mathbf{V}^{l}$ is constructed through three successive operations. First, patch-aware refinement adapts learnable micro-forensic primitives to the input patches, enabling them to capture reusable and composable fine-grained evidence units. Second, class-specific global contextual prompts provide routing signals to dynamically select and compose the refined primitives into real and spoof-specific primitive evidence prompts. Finally, the resulting primitive evidence prompts are augmented with the corresponding global contextual prompts to form $\mathbf{V}^{l}$. The following subsections detail the patch-aware primitive refinement and globally guided primitive composition mechanisms.

\subsection{Patch-Aware Micro-Forensic Primitive Refinement}
Spoofing cues in open-world face anti-spoofing are highly heterogeneous in both appearance and spatial distribution. Across and within attack types, variations in materials, structures, and fabrication processes produce diverse geometric, reflective, textural, and boundary artifacts. These cues may be global or localized, salient or subtle, making attack-specific templates prone to overfitting seen categories.
Nevertheless, unseen attacks often represent novel combinations of recurring forensic evidence rather than entirely new physical mechanisms. This motivates learning reusable forensic primitives and adaptively composing them for each input, offering a more generalizable alternative to attack-type modeling.

Accordingly, we introduce micro-forensic primitives to organize heterogeneous spoofing patterns into reusable and composable evidence units. At stage $l$, the learnable primitive set is defined as
\[
\mathbf{P}^{l}
=
\left\{
p_{i}^{l}
\right\}_{i=1}^{N_{P}^{l}}
\in
\mathbb{R}^{N_{P}^{l}\times d},
\]
where \(N_{P}^{l}\) is the number of primitives and \(d\) is the token dimension. Each primitive serves as an elementary evidence probe for localized, fine-grained, and high-frequency forensic cues, such as specular anomalies, boundary discontinuities, and local texture absence.

Because such evidence is often sparsely distributed over local patches, we introduce a patch-aware attention mechanism to refine each primitive using input-dependent local evidence. 
Given an input image $\mathbf{x}$, the patch embedding layer converts it into an initial token sequence:
$
\mathbf{T}^0 = [\mathbf{x}_{cls}^0, \mathbf{x}_{1}^0, \mathbf{x}_{2}^0, \cdots, \mathbf{x}_{N_t}^0],
$
where $\mathbf{x}_{cls}^0$ is the classification token, and $\{\mathbf{x}_{i}\}_{i=1}^{N_t}$ are the $N_t$ patch tokens.
Let
$
\mathbf{T}^{l}
=
[\mathbf{x}_{1}^{l},
\mathbf{x}_{2}^{l},
\ldots,
\mathbf{x}_{N_t}^{l}]
$
denote the patch tokens at stage \(l\). Treating the primitives as queries and the patch tokens as keys and values, we compute the attention weight between the \(i\)-th primitive and the \(j\)-th patch as
\begin{equation}\label{equ5}
\begin{aligned}
	a_{i,j}^{l} =
	\frac{
		\exp\left((q_i^l)^\top k_j^l / \sqrt{d}\right)
	}{
		\sum_{t=1}^{N_t}
		\exp\left((q_i^l)^\top k_t^l / \sqrt{d}\right)
	},
\end{aligned}
\end{equation}
where
$
q_i^l=p_i^l\mathbf{W}_q,\quad
k_j^l=\mathbf{x}_j^l\mathbf{W}_k,\quad
v_j^l=\mathbf{x}_j^l\mathbf{W}_v.
$
Here, \(\mathbf{W}_{q}\), \(\mathbf{W}_{k}\), and \(\mathbf{W}_{v}\) are learnable projection matrices. The coefficient \(a_{i,j}^{l}\) measures how strongly the \(i\)-th primitive responds to the \(j\)-th patch, allowing each primitive to selectively capture regions containing relevant forensic evidence.

The refined primitive is then obtained by aggregating the corresponding local evidence as
\begin{equation}\label{equ6}
	\begin{aligned}
\hat{p}_i^l=\sum_{j=1}^{N_t}a_{i,j}^{l}v_j^l.
	\end{aligned}
\end{equation}
Micro-forensic primitives are implemented as learnable vectors shared across all training samples and presentation categories. Through patch-aware attention, each primitive acts as a query to selectively retrieve and aggregate relevant evidence from local patches, encouraging it to capture localized physical patterns. During joint optimization over diverse categories and samples, recurring local patterns provide more consistent learning signals than incidental factors such as background variation and sample-specific noise. This shared learning process encourages individual primitives to develop preferences for different types of local evidence and promote complementary functional roles among them. Since the primitives represent shared candidate evidence rather than category-specific templates, a primitive can potentially contribute to both real and heterogeneous spoof representations, facilitating reuse across presentation categories. Such specialization and reusability are not explicitly enforced by the architecture, but are instead encouraged by the inductive biases introduced through shared parameterization, patch-aware evidence aggregation, and joint optimization across categories.

Because different stages of the vision foundation model encode distinct abstraction levels, we assign an independent primitive set to each stage. This design aligns primitive learning with the hierarchical feature space and enables multi-scale forensic modeling. Overall, patch-aware refinement avoids reliance on attack-specific semantics or fixed category templates and yields a shared library of input-refined visual evidence units for subsequent class-conditioned composition.


\subsection{Globally Guided Primitive Routing and Compositional Inference}

A presentation attack is typically characterized by multiple forensic cues, while different attacks exhibit distinct combinations of such cues. Since the primitive pool is shared across samples and presentation categories, the relevance of each primitive varies with the input: the same primitive may provide useful evidence for multiple categories, whereas different inputs require different combinations and weighting of primitives. Therefore, treating all primitives equally may dilute discriminative cues and introduce irrelevant responses, motivating input-adaptive primitive composition conditioned on global context.

We therefore use global context to condition primitive routing and composition. The global contextual prompts interact with the image tokens through the frozen backbone and are intended to summarize image-level context relevant to the real/spoof decision, including acquisition and presentation characteristics. They then provide sample-specific and class-specific conditions for weighting local evidence. At stage \(l\), the global contextual prompt is defined as
\begin{equation}\label{equ227}
	\mathbf{G}^{l}
	=
	[
	\mathbf{g}_{\mathrm{real}}^{l},
	\mathbf{g}_{\mathrm{spoof}}^{l}
	],
\end{equation}
where the two tokens correspond to the real and spoof classes, respectively. The global contextual prompt is initialized at the beginning of training, inserted after the \(\mathrm{CLS}\) token and before the patch tokens, and propagated through successive backbone stages. Through self-attention, it interacts with both global and local image representations. Accordingly, \(\mathbf{G}^{l}\) denotes its representation at stage \(l\), rather than an independently initialized stage-specific prompt.

A lightweight routing network maps each class-specific contextual token to a distribution over the refined primitives as
\begin{equation}\label{equ7}
	\boldsymbol{\beta}^{l}
	=
	[
	\boldsymbol{\beta}_{\mathrm{real}}^{l};
	\boldsymbol{\beta}_{\mathrm{spoof}}^{l}
	]
	\in
	\mathbb{R}^{2\times N_{P}^{l}},
\end{equation}
where
\begin{equation}\label{equ8}
	\boldsymbol{\beta}_{c}^{l}
	=
	\operatorname{Softmax}
	\left(
	\operatorname{MLP}(\mathbf{g}_{c}^{l})
	\right),
	\qquad
	c\in\{\mathrm{real},\mathrm{spoof}\}.
\end{equation}
The softmax operation is applied along the primitive dimension
\begin{equation}\label{equ9}
	\boldsymbol{\beta}_{c}^{l}
	=
	[
	\beta_{c,1}^{l},
	\beta_{c,2}^{l},
	\ldots,
	\beta_{c,N_{P}^{l}}^{l}
	],
	\qquad
	\sum_{i=1}^{N_{P}^{l}}
	\beta_{c,i}^{l}
	=
	1,
\end{equation}
where \(\beta_{c,i}^{l}\) quantifies the contribution of the \(i\)-th primitive to the evidence representation of class \(c\). Each global contextual token therefore acts as a class-specific router that modulates local forensic evidence according to the holistic physical characteristics of the input.
Given the refined primitive set
$
\hat{\mathbf{P}}^{l}
=
\left\{
\hat{p}_{i}^{l}
\right\}_{i=1}^{N_{P}^{l}},
$
the class-specific primitive evidence prompt is constructed as
\begin{equation}\label{equ10}
	\mathbf{c}_{c}^{l}
	=
	\sum_{i=1}^{N_{P}^{l}}
	\beta_{c,i}^{l}
	\hat{p}_{i}^{l},
	\qquad
	c\in\{\mathrm{real},\mathrm{spoof}\}.
\end{equation}
The resulting prompts are represented as
\begin{equation}\label{equ11}
	\mathbf{C}^{l}
	=
	[
	\mathbf{c}_{\mathrm{real}}^{l},
	\mathbf{c}_{\mathrm{spoof}}^{l}
	].
\end{equation}
The refined primitives summarize candidate local evidence, while the global contextual prompts determine their class-specific routing weights for the current input. Class-specific routing therefore forms distinct real- and spoof-oriented combinations from the same primitive pool. This global-to-local conditioning provides an explicit mechanism for suppressing weakly weighted responses and constructing input-adaptive evidence combinations for both seen and unseen attack types.

Finally, the dynamically composed primitive evidence prompt is further augmented with the corresponding global contextual prompt to obtain the compositional forensic visual prompt \(\mathbf{V}^{l}\) for the stage \(l\) as
\begin{equation}\label{equ1}
	\mathbf{V}^{l}
	=
	\mathbf{G}^{l}
	+
	\mathbf{C}^{l}.
\end{equation}
The resulting stage-specific prompt preserves the composition of localized forensic evidence while incorporating complementary image-level context.

\subsection{Compositional Prompt Aggregation and Optimization}
The stage-specific compositional forensic visual prompts are aggregated across representation depths to obtain the final prompt as
\begin{equation}\label{equ2}
	\mathbf{V}^{*}
	=
	\sum_{l=1}^{L}\mathbf{V}^{l},
\end{equation}
where
$
\mathbf{V}^{*}
=
[
\mathbf{v}_{\mathrm{real}}^{*},
\mathbf{v}_{\mathrm{spoof}}^{*}
]
$
contains the final class-specific prompt representations. For each class \(c\), the classification logit is computed by measuring the similarity between its prompt representation and the final \(\mathrm{CLS}\) embedding  \(\mathbf{z}_{\mathrm{cls}}\) as
\begin{equation}\label{equ3}
	o_{c}
	=
	\operatorname{sim}
	(
	\mathbf{z}_{\mathrm{cls}},
	\mathbf{v}_{c}^{*}
	),
	\qquad
	c\in\{\mathrm{real},\mathrm{spoof}\},
\end{equation}
where \(\operatorname{sim}(\cdot,\cdot)\) denotes the similarity function. To support unified discrimination across heterogeneous attacks, all presentation attack types share the binary label space
$
\mathcal{C}=\{\mathrm{real},\mathrm{spoof}\},
$
where both seen and unseen attacks are assigned to the \(\mathrm{spoof}\) class.
The model is optimized using the cross-entropy loss as
\begin{equation}\label{equ4}
	\mathcal{L}_{\mathrm{ce}}
	=
	-
	\sum_{c\in\mathcal{C}}
	y_c
	\log
	\frac{\exp(o_c)}
	{\sum_{k\in\mathcal{C}}\exp(o_k)},
\end{equation}
where \(y_c\) represents the one-hot ground-truth label. 

During training, the vision foundation model remains frozen, and only the global contextual prompts, micro-forensic primitives, projection layers, and lightweight routing modules are optimized. This parameter-efficient learning strategy preserves the general visual knowledge of the pretrained backbone while adapting it to compositional forensic diagnosis.
During inference, given an input image $x$, the framework first constructs the corresponding compositional forensic visual prompts, computes the class logits using Eq.~\ref{equ3}, and applies softmax normalization to obtain the prediction probabilities.

Overall, the proposed framework performs prompt learning entirely in the continuous visual feature space. By dynamically combining refined primitive queries across hierarchical representation levels, it provides an input-adaptive representation that spans image-level context and localized evidence without requiring predefined attack-category descriptions. In this sense, the method implements compositional inference through explicit weighted combinations of shared visual evidence units for both seen and unseen attack types.

\section{Experiments}
\label{sec4}
\subsection{Experimental Setups}
\label{sec41}
\noindent \textbf{Datasets.}
We conduct experiments on eight public benchmarks: CASIA-MFSD\cite{zhang2012face} (C), Replay-Attack\cite{chingovska2012effectiveness} (I), MSU-MFSD\cite{wen2015face} (M), OULU-NPU\cite{boulkenafet2017oulu} (O), HQ-WMCA\cite{9146362hqwmca} (H), SiW-Mv2\cite{guo2022multi} (W), CASIA-SURF\cite{zhang2020casia} (S), and CASIA-SURF CeFA\cite{liu2021casia} (F). These datasets cover substantial variations in acquisition devices, illumination, backgrounds, attack types, providing a comprehensive evaluation of cross-domain and unseen-attack generalization in open-world scenarios.

CASIA-MFSD, Replay-Attack, MSU-MFSD, and OULU-NPU primarily contain conventional print and replay attacks. CASIA-MFSD includes warped-photo, cut-eye, and video-replay attacks captured under different imaging qualities. Replay-Attack contains photo and video attacks presented using printed media and electronic displays under both controlled and adverse illumination. MSU-MFSD introduces variations in acquisition devices and presentation media through printed photos and replay attacks generated by mobile devices. OULU-NPU targets mobile authentication scenarios and systematically varies smartphone cameras, illumination, backgrounds, printers, and display devices. Together, these datasets provide diverse yet relatively constrained source-domain attack distributions.

HQ-WMCA and SiW-Mv2 contain substantially more diverse attack types and are therefore used to evaluate generalization to heterogeneous unseen attacks. HQ-WMCA is a multimodal benchmark covering visible, depth, thermal infrared, near-infrared, and short-wave infrared data, with attacks including print, replay, rigid mask, paper mask, flexible mask, mannequin, glasses, makeup, tattoo, and wig. SiW-Mv2 contains 14 attack types, spanning conventional print and replay attacks, as well as 
half mask, paper mask, transparent mask, silicone mask, mannequin, partial eye, funny eye glasses, partial mouth, paper glasses, cosmetic makeup, impersonation makeup, obfuscation makeup. Both datasets exhibit pronounced heterogeneity in attack appearance, physical properties, and spatial distributions.

CASIA-SURF is a large-scale multimodal dataset providing visible, depth, and near-infrared observations for each sample, with print-based presentation attacks. CASIA-SURF CeFA further introduces cross-ethnicity, multimodal, and attack-type variations, covering Asian, African, and Central Asian subjects as well as print, replay, 3D-printed mask, and silicone-mask attacks. The CASIA-SURF to CeFA setting therefore provides a large-scale benchmark involving both covariate and semantic shifts. Since this work focuses on visible-spectrum face anti-spoofing, only the visible modality is used for all multimodal datasets.

\noindent \textbf{Evaluation protocols and metrics.}
We construct nine cross-dataset open-world protocols. Eight protocols are formed by selecting one source domain from CASIA-MFSD, Replay-Attack, MSU-MFSD, and OULU-NPU and one target domain from HQ-WMCA and SiW-Mv2. The source datasets contain print and replay attacks, whereas the target datasets include substantially more diverse attack categories. We additionally evaluate the protocol from CASIA-SURF to CASIA-SURF CeFA to assess open-world generalization at a larger scale.
These protocols introduce simultaneous discrepancies in acquisition devices, imaging conditions, and attack semantics, thereby jointly evaluating robustness to covariate and semantic shifts. Half Total Error Rate (HTER) and Area Under the ROC Curve (AUC) are adopted as the evaluation metrics.

\noindent \textbf{Implementation details.}
The proposed method is implemented in PyTorch. We employ the pretrained CLIP ViT-L/14@336px image encoder as a frozen backbone and insert the proposed visual prompt modules at layers 6, 12, 18, and 24. Input images are resized to $259\times259$, and the pretrained positional embeddings are bilinearly interpolated to match the resulting token grid. Eight micro-forensic primitives are used at each stage, with a batch size of 32. The model is optimized using AdamW with $\beta_1=0.9$, $\beta_2=0.999$, a weight decay of 0.01, and an initial learning rate of $1\times10^{-3}$, which is decayed using a cosine annealing schedule over 30 training epochs. For preprocessing, face regions are detected and cropped from video frames using the method in~\cite{Zhang2016Joint} and stored at a resolution of $128\times128$. During training, random horizontal flipping with a probability of 0.5 is applied for data augmentation.

\begin{figure*}[htbp]
	\centering
	\includegraphics[width=1\textwidth]{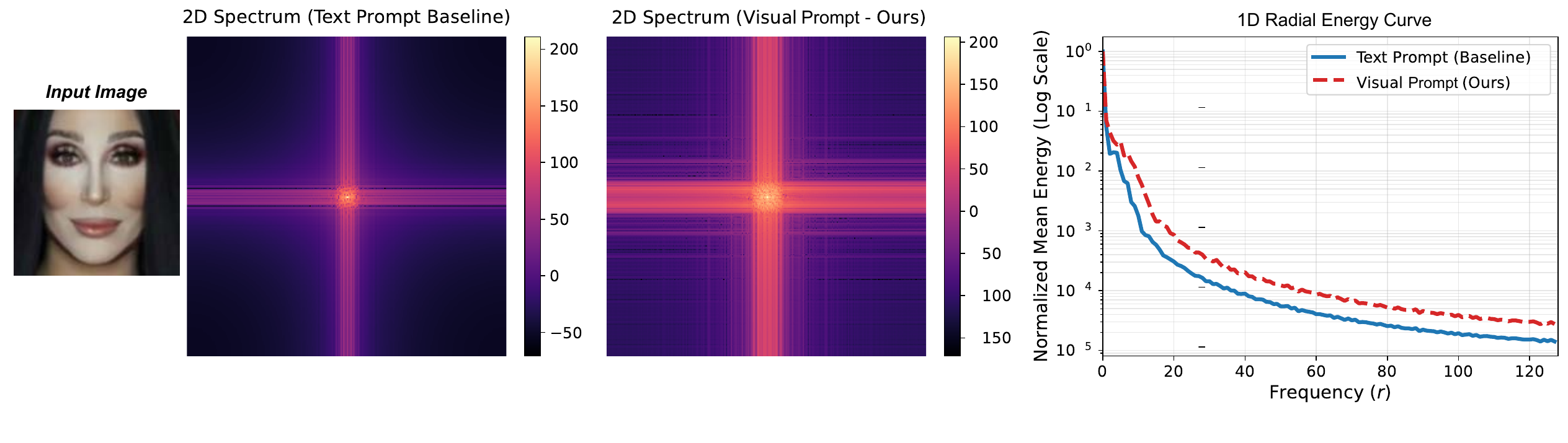}
	\caption{
		Fourier analysis results of the text-prompt baseline and the proposed visual prompts under the $C \;to\; W$ protocol. 
		The 2D spectrum show that both prompt types produce strong low-frequency responses, whereas the proposed visual prompts exhibit more pronounced energy expansion in the high-frequency regions. The 1D radial energy curves further demonstrate that visual prompts maintain stronger and more stable responses in the mid- and high-frequency ranges, indicating their superior ability to capture fine-grained forensic cues and thereby enhance cross-domain unseen-attack detection.
}
	\label{fourieranalysis}
\end{figure*}

\subsection{Visualization Analysis of Compositional Forensic Visual Prompts}
To comprehensively examine the learned compositional forensic visual prompts, we conduct visualization analyses from four perspectives. We first compare visual and textual prompts in the frequency domain to evaluate their ability to capture fine-grained spoofing artifacts, particularly high-frequency forensic cues. We then visualize the spatial responses of compositional forensic visual prompts to investigate their localization of discriminative forensic evidence. Next, layer-wise attention maps are analyzed to reveal how forensic evidence evolves across different representation levels. Finally, t-SNE visualization is employed to examine the distribution and discriminability of different prompt components in the feature space.

\subsubsection{Frequency-Domain Comparison with Text Prompts}
To investigate whether the proposed visual prompts capture richer high-frequency forensic evidence than text prompts, we conduct Fourier analysis under the $C \;to\; W$  protocol. CoOp\cite{Zhou_2022} is adopted as the text-prompt baseline and trained on the same source domain. For each target-domain sample, we compute the similarity between the learned prompts and the last-layer patch tokens to obtain a spatial activation map. We then calculate its two-dimensional (2D) Fourier spectrum and the corresponding one-dimensional (1D) radial energy curve, where the radius represents the spatial frequency.

As shown in \figurename~\ref{fourieranalysis}, both prompt types exhibit strong responses near the spectral center, indicating comparable sensitivity to low-frequency semantics and global structures. However, the visual prompts show a broader energy distribution toward the spectral periphery, suggesting that they activate richer high-frequency responses. The radial energy curves further reveal that, although both prompts retain strong low-frequency responses, the energy of text prompts decays rapidly with increasing frequency, whereas visual prompts preserve substantially stronger mid- and high-frequency responses. These results demonstrate that the proposed visual prompts are more sensitive to fine-grained high-frequency spoofing artifacts. They also validate the effectiveness of compositional forensic visual prompts in jointly modeling global low-frequency structures and local high-frequency forensic evidence.

\begin{figure}[tb]
	\centering
	\includegraphics[width=0.49\textwidth]{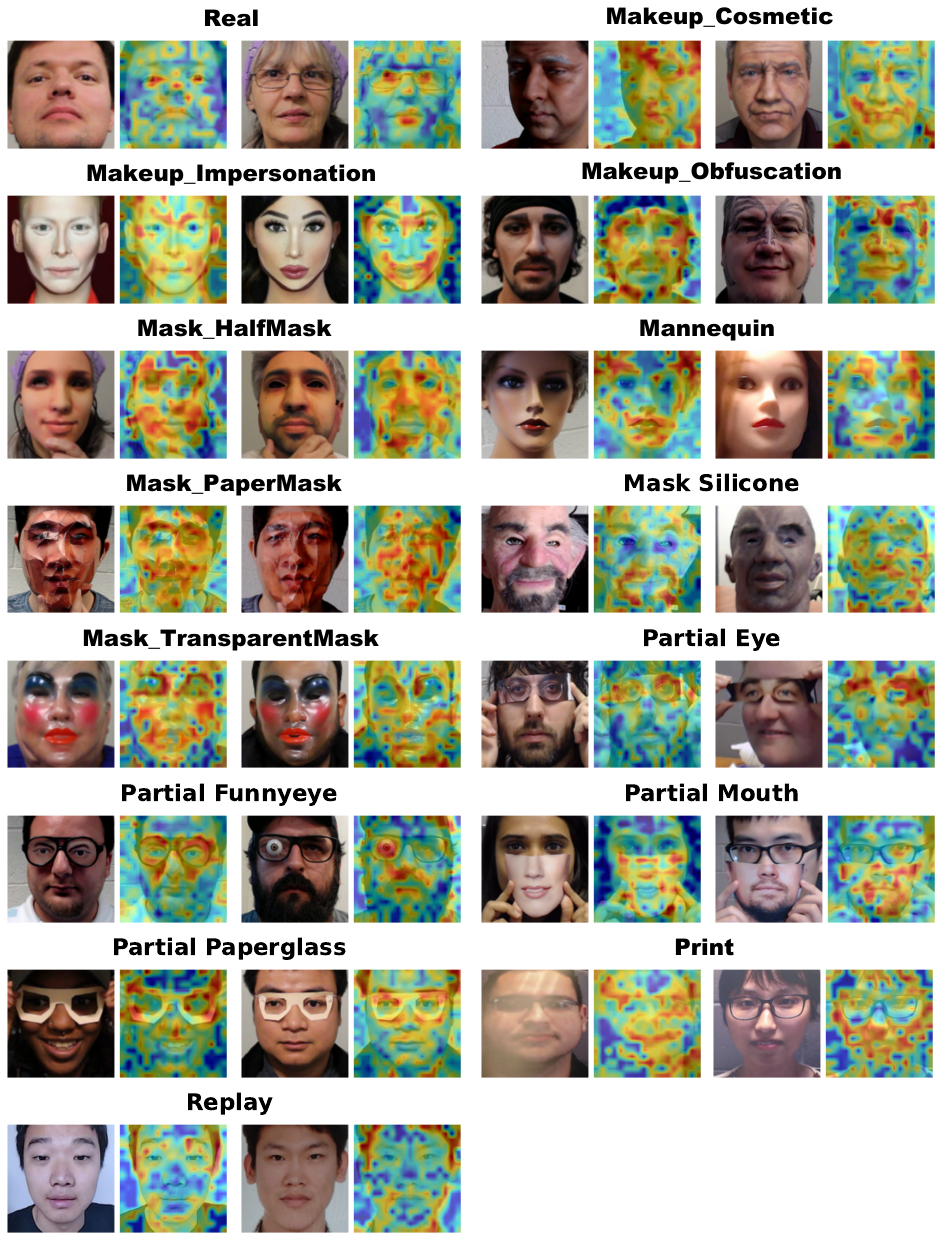}
	\caption{
		Visualization of attention regions activated by the compositional forensic visual prompts under the $C \;to\; W$ protocol. Real faces mainly activate structurally informative regions, such as the eyes, nose, and mouth, while spoof samples activate attack-specific forensic regions, including makeup artifacts, full-face mask areas, and localized eye- or mouth-level spoof traces. These results show that the learned prompts can effectively localize discriminative evidence across heterogeneous attacks.
	}
	\label{caml3}
\end{figure}

\subsubsection{Spatial Response Visualization of Compositional Forensic Visual Prompts}
To examine the spatial evidence captured by the learned prompts, we visualize their activation maps under the $C \;to\; W$ protocol. For each sample, the activation map is generated from the similarity between the learned prompts and the last-layer patch tokens. As shown in \figurename~\ref{caml3}, real samples mainly activate structurally informative regions, including the eyes, nose, and mouth. In contrast, spoof samples exhibit responses aligned with their attack-specific artifacts. Makeup attacks primarily activate regions with abnormal color, texture, wrinkles, or painted patterns; full-face masks produce broadly distributed responses over the facial area; and partial attacks concentrate on localized spoof regions, such as the eyes, funny-eye regions, or mouth. The consistent responses observed across samples of the same attack type qualitatively suggest that the learned prompts attend to spatially relevant evidence across heterogeneous attacks.

\subsubsection{Layer-Wise Visualization of Prompt Attention}
To examine the hierarchical diagnostic behavior of the proposed framework, we visualize the attention regions of visual prompts at different layers under the $C \;to\; W$ protocol, using partial-eye and partial-mouth attacks with clearly localized spoofing cues as example. For each sample, we compute the similarity between the layer-specific prompt representations and the corresponding patch tokens to generate spatial activation maps.

\begin{figure}[tb]
	\centering
	\includegraphics[width=0.48\textwidth]{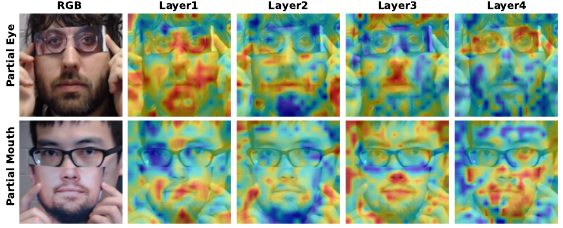}
	\caption{
		Visualization of attention regions across visual prompt layers, using partial eye and partial mouth attacks under the $C \;to\; W$ protocol.
		The results show a progressive shift from global facial context to localized spoofed regions, indicating that the learned prompts guide the frozen vision foundation model toward forensic-level evidence localization.
	}
	\label{cambianhua}
\end{figure}

As shown in \figurename~\ref{cambianhua}, the prompt responses progressively evolve from global to local and from coarse- to fine-grained patterns. At shallow layers, the prompts exhibit broad responses over the facial region and surrounding context, primarily capturing low-level appearance and global facial structure. At intermediate layers, the activations gradually concentrate on facial boundaries and geometrically informative regions, such as the forehead and nose, indicating increased sensitivity to structural and physical inconsistencies. At deeper layers, the responses further converge on the manipulated eye or mouth regions, enabling precise localization of fine-grained spoofing artifacts.

This hierarchical evolution indicates that the learned prompts progressively guide the frozen vision foundation model from generic visual perception toward forensic-level cues localization. By progressively refining attention across layers, the model gradually suppresses task-unrelated interference and locks onto local discriminative evidence associated with unseen attacks, thereby enhancing generalization in open-world scenarios.

\begin{figure}[htbp]
	\centering
	\subfloat[Global Contextual Prompts\label{fig:sub2}]{\includegraphics[width=0.47\textwidth]{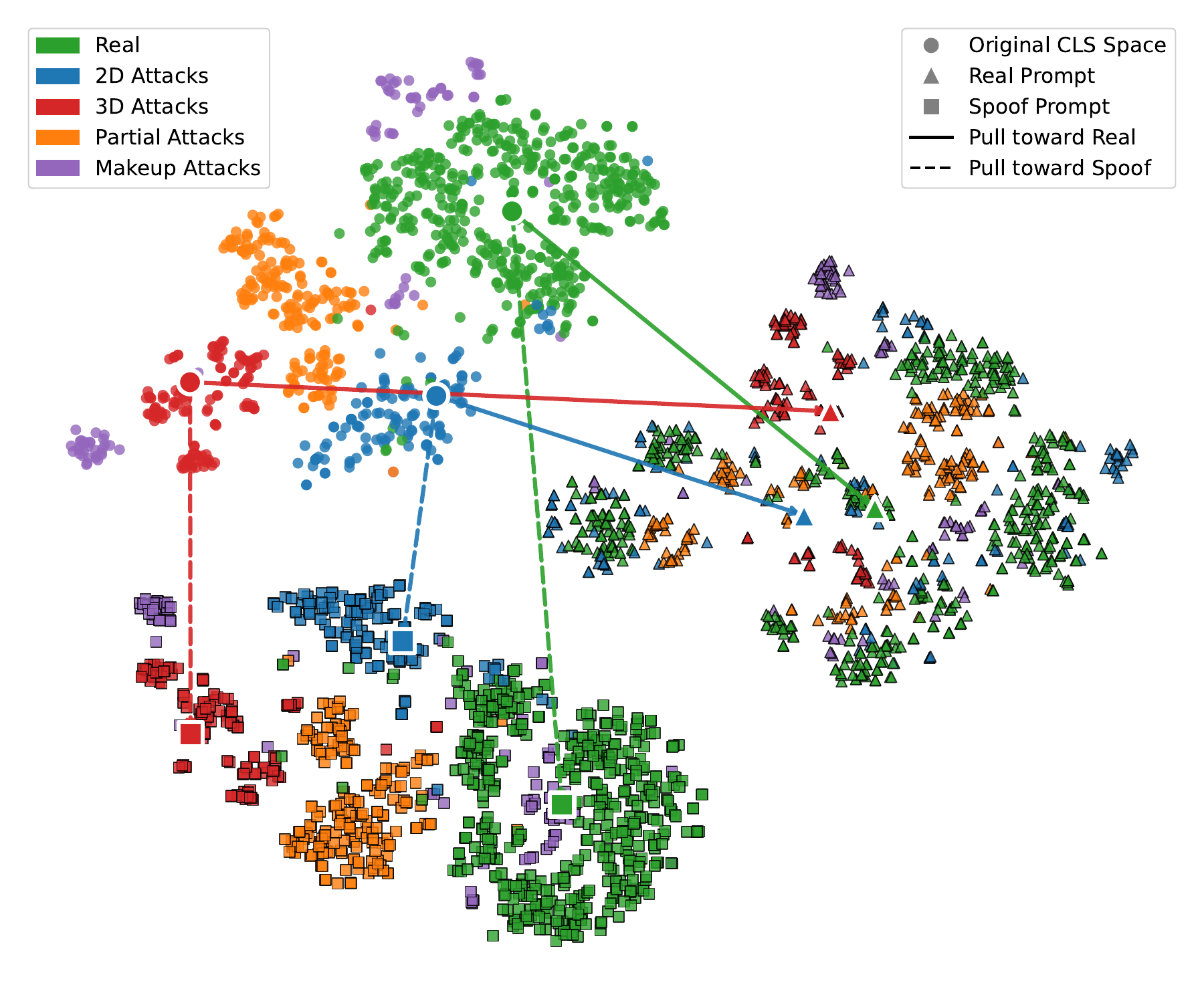}}%
	\hfill
	\subfloat[Primitive Evidence Prompts\label{fig:sub3}]{\includegraphics[width=0.47\textwidth]{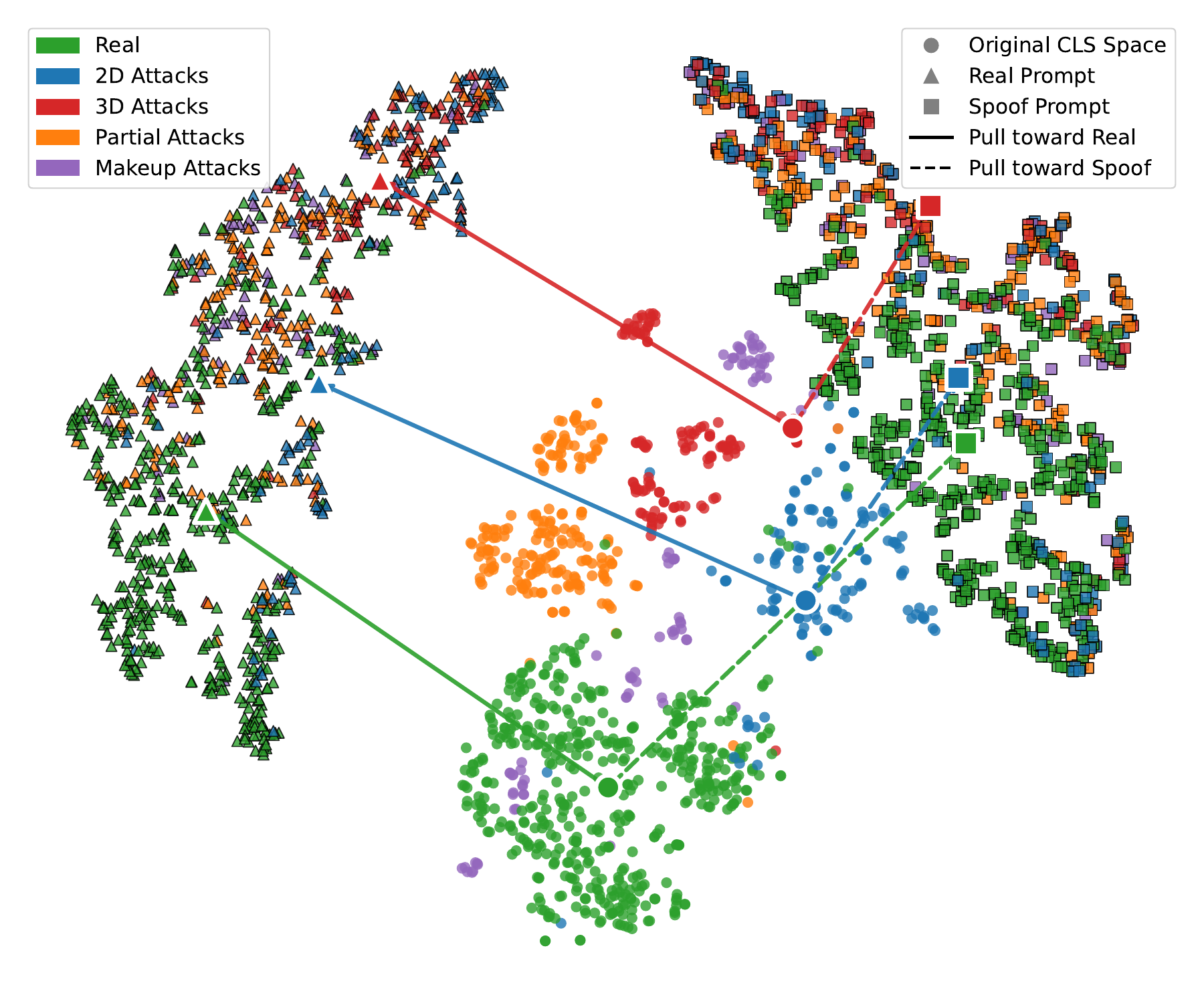}}%
	\hfill
	\subfloat[Compositional Forensic Visual Prompts\label{fig:sub1}]{\includegraphics[width=0.47\textwidth]{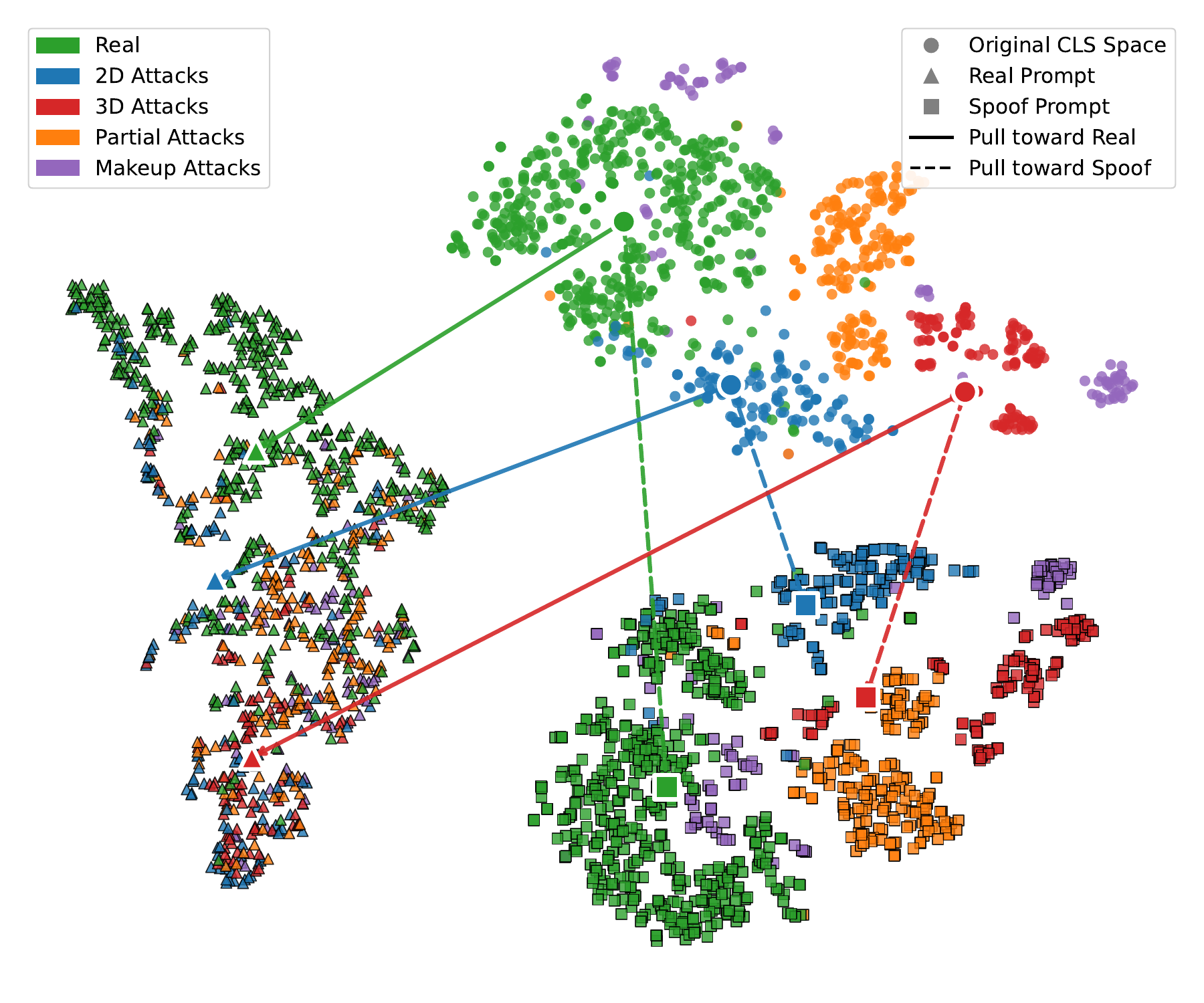}}%
	\caption{t-SNE visualization analysis of different prompt components.}
	\label{fig:threefigs}
\end{figure}
\subsubsection{t-SNE Analysis of Different Prompt Components}
To investigate the discriminative structures learned by different prompt components, we visualize the compositional forensic visual prompts, global contextual prompts, and primitive evidence prompts using t-SNE under the $C \;to\; W$ protocol. For clarity, the 14 target-domain attacks are grouped into four categories according to their physical characteristics: 2D attacks, including print and replay; 3D attacks, including half mask, paper mask, transparent mask, silicone mask, and mannequin; partial attacks, including partial eye, funny-eye glasses, partial mouth, and paper glasses; and makeup attacks, including cosmetic, impersonation, and obfuscation makeup. We randomly sample 1,000 target-domain instances, extract their class-specific prompt embeddings, and use category prototypes to denote the corresponding feature centroids.

As shown in \figurename~\ref{fig:threefigs}, all three types of prompts exhibit a certain degree of real/spoof discriminability. Real samples are generally closer to the real prompts, whereas spoof samples of different types are closer to the spoof prompts, indicating that the learned prompts can form discriminative directions in the feature space that are consistent with the classification objective. 
A further comparison shows that the real global contextual prompts of real samples are relatively dispersed, with evident overlap between real samples and 2D, 3D, and makeup attacks. Moreover, the spoof global contextual prompts of real samples surround those of makeup attacks. This suggests that relying solely on image-level global semantics remains vulnerable to interference from identity, imaging conditions, and appearance similarity.

In contrast, the primitive evidence prompts produce a clearer separation between real and spoof samples, indicating that micro-forensic primitives capture transferable local evidence, and the primitive evidence prompts achieve effective evidence composition. 
The compositional forensic visual prompts form a more compact distribution of real prompts and a clearer real/spoof decision boundary, while effectively alleviating the surrounding and entanglement phenomena observed in the global contextual prompts.
These results demonstrate that the proposed composition mechanism effectively integrates local forensic evidence and global physical priors, thereby improving prompt discriminability and generalization capability in open-world scenarios with heterogeneous attacks.

\begin{figure}[tb]
	\centering
	\includegraphics[width=0.5\textwidth]{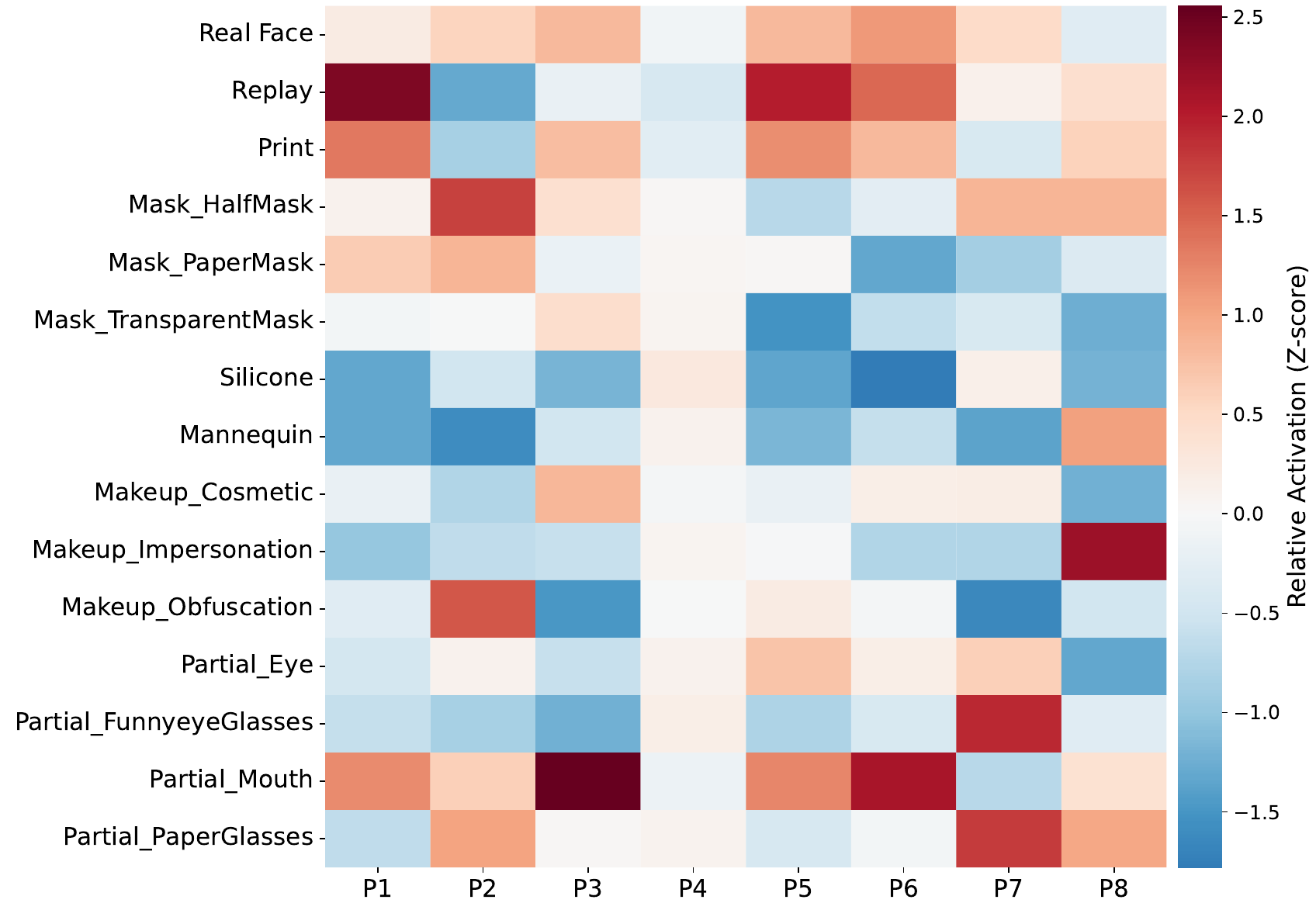}
	\caption{
		Category activation visualization of micro-forensic primitives under the $C \;to\; W$ protocol. Different primitives exhibit clear category selectivity and produce distinct responses to both real and spoof evidence. Meanwhile, different attack types share a subset of primitives, whereas a single primitive cannot cover all spoofing patterns, indicating that heterogeneous attacks should be jointly represented by multiple reusable primitives. These results demonstrate that micro-forensic primitives provide an effective foundation for dynamic compositional modeling of unseen attacks in open-world scenarios.
	}
	\label{prototype_heatmapl3}
\end{figure}

\subsection{Visualization Analysis of Micro-Forensic Primitives}
To further investigate the learned micro-forensic primitives, we conduct visualization analyses from three perspectives. We first analyze primitive activations across different presentation categories to reveal their category-dependent response patterns. We then examine how these activation patterns evolve across layers to characterize the progressive formation of forensic representations. Finally, we visualize the image patches most strongly associated with each primitive to investigate the localized visual evidence captured by different primitives.

\subsubsection{Primitive Category Activation Analysis}
To analyze the category selectivity of the learned primitives, we compute the average similarity between each primitive and all patch tokens from samples of a given category under the $C \;to\; W$ protocol. The responses of each primitive are then normalized across categories using z-score normalization, producing the activation map shown in \figurename~\ref{prototype_heatmapl3}.

Different primitives exhibit distinct category preferences. Some primitives show strong activation for print, replay, and partial mouth attacks, suggesting that they may capture shared evidence such as 2D medium artifacts, local texture degradation, and fine-grained spoofing traces. Some primitives respond more strongly to attacks such as paper glasses, funny eye glasses, half mask, paper mask, mannequin, and makeup, indicating greater sensitivity to forensic cues related to local occlusion, mask materials, geometric discontinuities, and appearance disguise.

Several primitives also exhibit strong responses to real faces, showing that the primitive library models not only spoofing artifacts but also normal structural and textural evidence. Moreover, different attacks share partially overlapping high-response primitives, while each attack is typically represented by multiple primitives. This observation supports the hypothesis that heterogeneous attacks can be represented by distinct combinations of shared evidence units. It also validates the effectiveness of the micro-forensic primitive library and further demonstrate the necessity of globally guided dynamic primitive composition for modeling unseen attacks in open-world scenarios.

\begin{figure}[tb]
	\centering
	\includegraphics[width=0.4\textwidth]{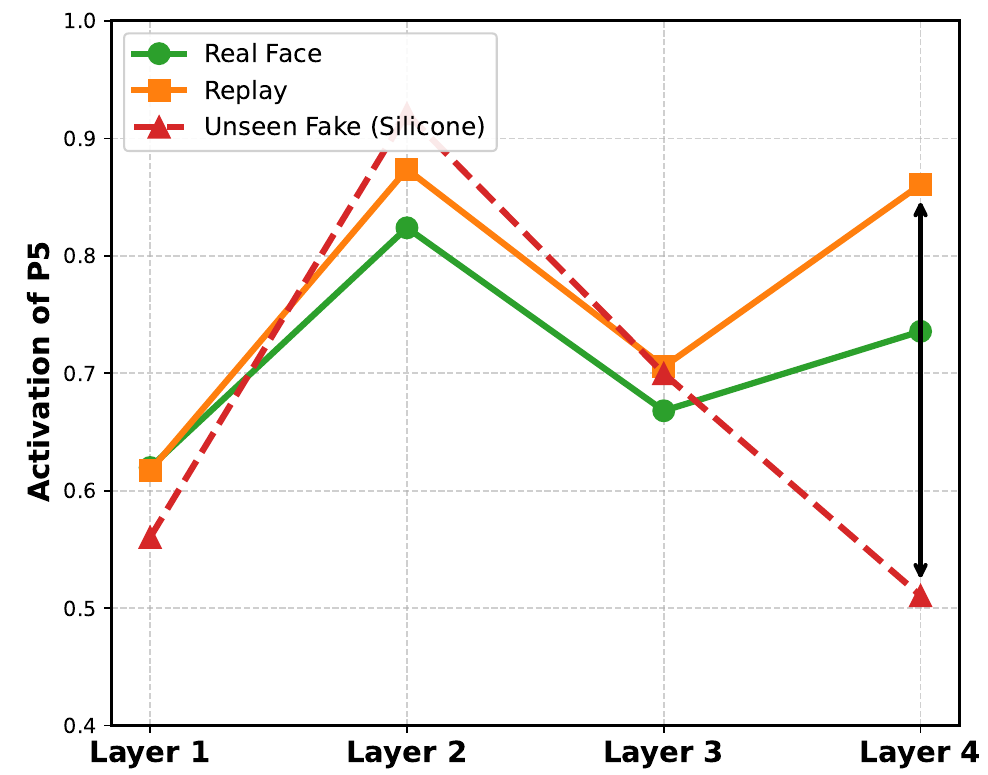}
	\caption{
		Evolution of category activations of primitive P5 across different layers under the $C \;to\; W$ protocol. Taking real faces, replay attacks, and silicone mask attacks as examples, the responses of P5 gradually evolve from cross-category similarity in shallow layers to clear category separation in deeper layers, indicating that micro-forensic primitives progressively develop stronger category selectivity and forensic discriminative capability as the network depth increases.
	}
	\label{figcrosslayerevolution}
\end{figure}

\subsubsection{Hierarchical Evolution of Primitive Category Activation}
To examine how primitive responses evolve across representation depths, we visualize the activation of primitive P5 for real faces, replay attacks, and silicone mask attacks under the $C \;to\; W$ protocol. As shown in \figurename~\ref{figcrosslayerevolution}, the three categories exhibit similar activations at Layer 1, indicating that the shallow representation of P5 mainly captures low-level patterns shared across categories and exhibits weak category selectivity.
As the network depth increases, the activation differences become progressively more pronounced. At Layer 2, P5 begins to show clearer category-specific response differences. At Layer 3, the activation values of replay attacks and silicone mask attacks are relatively close to each other, yet both are clearly separated from real faces, suggesting that at this stage P5 has acquired a certain capability to distinguish real from spoof samples, although its ability to differentiate between spoof types remains limited. At Layer 4, all three categories are more distinctly separated, indicating stronger category selectivity and improved sensitivity to heterogeneous forensic patterns. 
These results show that micro-forensic primitives progressively evolve from generic visual responses into category-relevant forensic cues, validating the effectiveness of multi-stage primitive learning for open-world face anti-spoofing.

\subsubsection{Visualization of Primitive-Related Patches}
To further understand the local evidence represented by each micro-forensic primitive, we retrieve and visualize its top-10 most similar patches under the $C \;to\; W$ protocol. Patch features are extracted from the final layer of the image encoder. To prevent the retrieval results from being dominated by a single sample, the top-10 patches for each primitive are constrained to originate from different images. Because the original patches are small, we additionally display their local context, with red boxes marking the matched regions.

\begin{figure}[tb]
	\centering
	\includegraphics[width=0.48\textwidth]{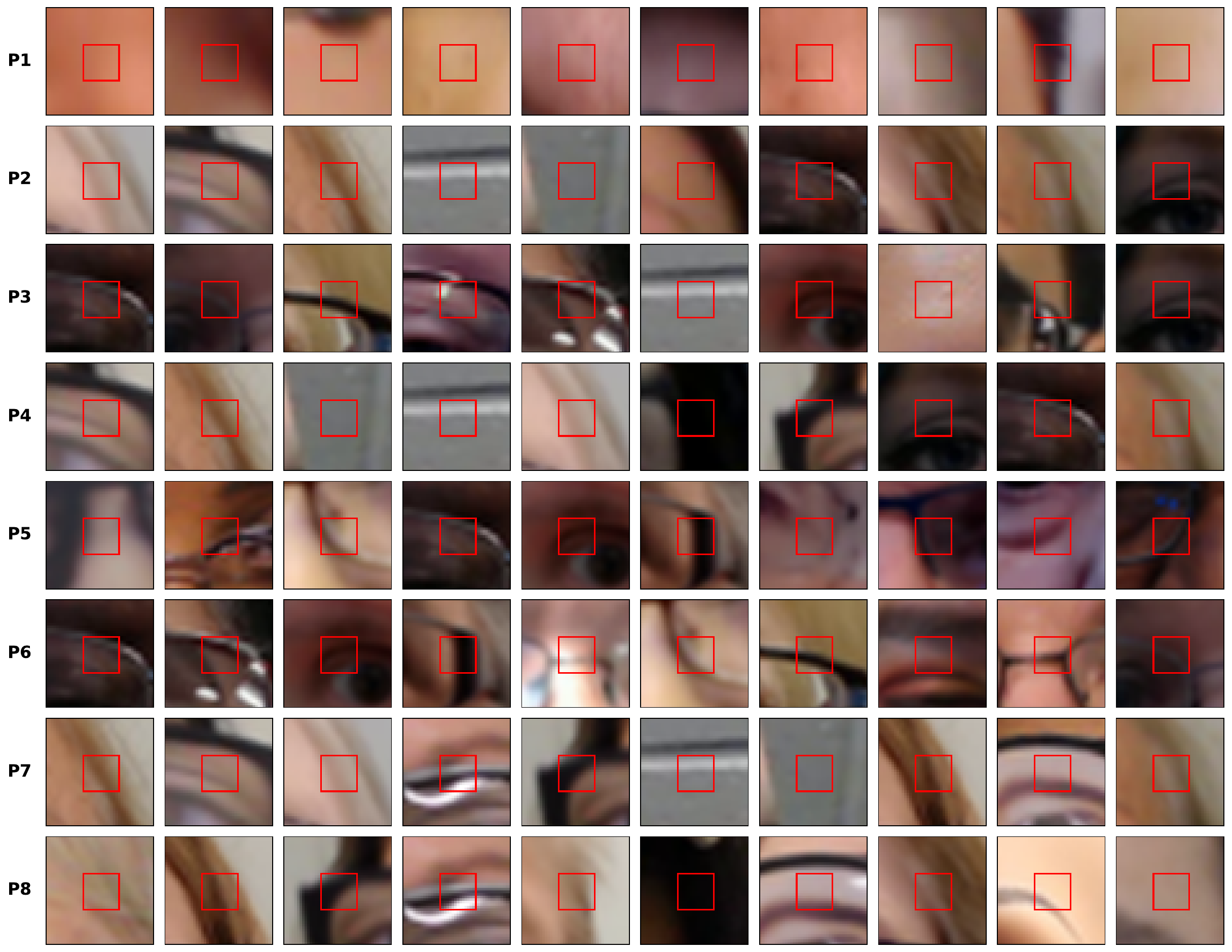}
	\caption{Visualization of the top-10 similar patches for micro-forensic primitives under the $C \;to\; W$ protocol. Red boxes mark the matched patch regions, and surrounding crops provide local context. 
	The retrieved examples show recurring local patterns for individual primitives and different preferences across primitives, providing qualitative evidence of specialization and reuse.	
	}
	\label{figtopkpatchesl3}
\end{figure}

As shown in \figurename~\ref{figtopkpatchesl3}, the same primitive consistently responds to similar local patterns across different samples, such as skin textures, eye and mouth structures, boundary transitions, and reflections. Different primitives exhibit distinct visual preferences, demonstrating functional specialization and complementarity within the primitive library. These results indicate that the primitives do not encode attack templates; instead, they represent transferable local forensic units shared across samples and categories, providing the basis for compositional modeling of heterogeneous attacks.

\begin{table*} 
	\small
	\caption{Comparison with state-of-the-art methods using SiW-Mv2 (W for short) dataset as the target domain, and one of CASIA-MFSD (C), Replay-Attack (I), MSU-MFSD (M), and OULU-NPU (O) dataset as the source domain. AVG represents the average value across the four protocols. A lower HTER indicates better performance, while a higher AUC signifies better performance.}
	\label{tabw}
	\centering
	\rowcolors{2}{white}{blue!5}
	\begin{tabular}{p{2.3cm}p{1.1cm}<{\centering}p{1.1cm}<{\centering}p{1.1cm}<{\centering}p{1.1cm}<{\centering}p{1.1cm}<{\centering}p{1.1cm}<{\centering}
			p{1.1cm}<{\centering}p{1.1cm}<{\centering}p{1.1cm}<{\centering}p{1.1cm}<{\centering}}
		\hline
		Method& \multicolumn{2}{c}{$C \;to\; W$} & \multicolumn{2}{c}{$I \;to\; W$} &\multicolumn{2}{c}{$M \;to\; W$}& \multicolumn{2}{c}{$O \;to\; W$}& \multicolumn{2}{c}{$AVG$}\\
		& HTER (\%)&AUC (\%) &HTER (\%)&AUC (\%)& HTER (\%)&AUC (\%)& HTER (\%)&AUC (\%)& HTER (\%)&AUC (\%)  \\
		MS-LBP \cite{maatta2011face} &43.84 &52.71& 54.12&44.42&54.19&46.15&42.53 &54.86 & 48.67 & 49.54\\
		Color texture  \cite{boulkenafet2016face} &45.10 &54.80&57.47 &41.80&46.84 &53.97& 47.17&53.55 & 49.15 & 51.03\\
		CNN	 \cite{yang2014learn} &41.92 &59.64&43.29 &53.84&43.56 &57.22&42.06 &57.88 & 42.71 & 57.15\\
		DTN\cite{liu2019deep}  & 45.39&55.13&47.61 &52.90&38.62 &65.61& 42.78&61.49 & 43.60 & 58.78\\
		SDTN\cite{liu2020disentangling}  &40.73 &61.89&42.91 &58.21&38.14 &63.97&46.39 &54.33 & 42.04 & 59.60\\
		DR-UDA\cite{wang2020unsupervised}  &45.08& 57.43&50.22& 49.77&49.18& 51.25&44.35& 58.93 & 47.21 & 54.35\\
		USDAN\cite{jia2021unified}  &35.05& 69.57& 38.64& 64.97&37.76& 64.39&35.52 &69.19 & 36.74 & 67.03\\
		PatchCNN\cite{wang2022patchnet}  &35.53& 67.93&46.52& 53.86&34.71& 68.21&35.74& 69.13 & 38.13 & 64.78\\
		PDEN\cite{li2021progressive}  &34.41 &70.15& 45.62&55.45&44.61 &56.65&44.28 &57.43 & 42.23 & 59.92\\
		LD\cite{wang2021learning}  & 33.96&71.01&41.37 &62.11&39.07 &62.96&36.08 &64.92 & 37.62 & 65.25\\
		IADG\cite{zhou2023instance}&49.11&50.76&45.82&52.82&48.52&49.19&46.72&56.27 & 47.54 & 52.26\\
		OSDG\cite{jiang2024open}  &31.96 &72.28&37.50 &65.51&33.24 &72.10&35.20& 69.28 & 34.48 & 69.79\\
		FoundPAD\cite{ozgur2025foundpad} &45.08&56.28&51.12&48.86&51.55&48.37&48.38&52.65 & 49.03 & 51.54\\
		MEFAS\cite{cai2025me}&13.65 &  94.90&41.16 &  63.69&35.15 &  70.92 &36.35&   68.33 & 31.58 & 74.46\\ 
		MVPFAS\cite{yu2025multi}&27.24&81.72&32.41 &74.97&54.63 &43.77&26.65& \textbf{79.92} & 35.23 & 70.10\\ 
		Ours&\textbf{10.14}&\textbf{95.26}&\textbf{29.40}&\textbf{75.99}&\textbf{23.86}&\textbf{84.21}&\textbf{25.63}&78.31 & \textbf{22.26} & \textbf{83.44}\\
		\hline
		
	\end{tabular}
\end{table*}

\begin{table*}  
	\small
	\caption{Comparison with state-of-the-art methods using HQ-WMCA (H for short) dataset as the target domain, and one of CASIA-MFSD (C), Replay-Attack (I), MSU-MFSD (M), and OULU-NPU (O) dataset as the source domain. AVG represents the average value across the four protocols. A lower HTER indicates better performance, while a higher AUC signifies better performance.
	}
	\label{tabh}
	\centering
	\rowcolors{2}{white}{blue!5}
	\begin{tabular}{p{2.3cm}p{1.1cm}<{\centering}p{1.1cm}<{\centering}p{1.1cm}<{\centering}p{1.1cm}<{\centering}p{1.1cm}<{\centering}p{1.1cm}<{\centering}
			p{1.1cm}<{\centering}p{1.1cm}<{\centering}p{1.1cm}<{\centering}p{1.1cm}<{\centering}}
		\hline
		Method& \multicolumn{2}{c}{$C \;to\; H$} & \multicolumn{2}{c}{$I \;to\; H$} &\multicolumn{2}{c}{$M \;to\; H$}& \multicolumn{2}{c}{$O \;to\; H$}& \multicolumn{2}{c}{$AVG$}\\
		& HTER (\%)&AUC (\%) &HTER (\%)&AUC (\%)& HTER (\%)&AUC (\%)& HTER (\%)&AUC (\%)& HTER (\%)&AUC (\%)  \\
		MS-LBP \cite{maatta2011face} &46.78 &56.63&53.85 &44.26& 50.52&51.45&43.24 &59.42 & 48.60 & 52.94 \\
		Color texture  \cite{boulkenafet2016face} &43.66 &61.70&49.63 &50.00&50.00 &50.00&42.20 &60.07 & 46.37 & 55.44 \\
		CNN	 \cite{yang2014learn} & 40.95&64.67&40.33&66.27&42.31 &63.87& 39.04&65.36 & 40.66 & 65.04 \\
		DTN\cite{liu2019deep}  & 39.04&62.86& 36.59&67.33&40.54 &65.22&45.64 &53.70 & 40.45 & 62.28 \\
		SDTN\cite{liu2020disentangling}  & 51.16&48.33&45.27 &56.35& 50.76&47.22& 41.92&62.07 & 47.28 & 53.49 \\
		DR-UDA\cite{wang2020unsupervised}  &40.43 &62.23&43.84 &57.19&42.70 &59.40& 39.82&64.09 & 41.70 & 60.73 \\
		USDAN\cite{jia2021unified}  &38.96 &66.30& 38.30&66.40&40.64 &62.80& 29.09 &75.60 & 36.75 & 67.78 \\
		PatchCNN\cite{wang2022patchnet}  &39.54 &64.54&35.03 &73.24&38.21 &67.28& 34.24 &65.77 & 36.76 & 67.71 \\
		PDEN\cite{li2021progressive}  &35.76 &69.10&31.03 &74.35&51.56 &48.48& 42.41&59.00 & 40.19 & 62.73 \\
		LD\cite{wang2021learning}  &38.33 &65.12&32.43 &73.57&42.87 &60.13&29.73 &76.38 & 35.84 & 68.80 \\
		IADG\cite{zhou2023instance}&45.91&54.19&48.75&51.08&45.30&57.02&42.95&59.90 & 45.73 & 55.55 \\ 
		OSDG\cite{jiang2024open}  &27.24 &78.81&28.03 &79.37&34.79 &71.26&27.23 &80.75 & 29.32 & 77.55 \\
		FoundPAD\cite{ozgur2025foundpad} &47.07&53.02&45.71&53.95&52.76&46.12&48.24&51.94 & 48.45 & 51.26 \\
		MEFAS\cite{cai2025me}&16.38&\textbf{90.43}&32.14 &  73.58&43.18 &  61.25&32.12  & 76.38 & 30.96 & 75.41 \\ 
		MVPFAS\cite{yu2025multi}&17.88&90.21&24.56&\textbf{85.16}&40.13&64.10&23.95&84.31 & 26.63 & 80.95 \\ 
		Ours&\textbf{15.11}&90.06&\textbf{23.63}&82.67&\textbf{28.97}&\textbf{76.61}&\textbf{22.13}&\textbf{84.84} & \textbf{22.46} & \textbf{83.55} \\
		\hline
	\end{tabular}
\end{table*}

\subsection{Quantitative Analysis of Primitive Routing}
\subsubsection{Primitive Similarity Analysis}

\begin{figure}[htpb]
	\centering
	\includegraphics[width=0.5\textwidth]{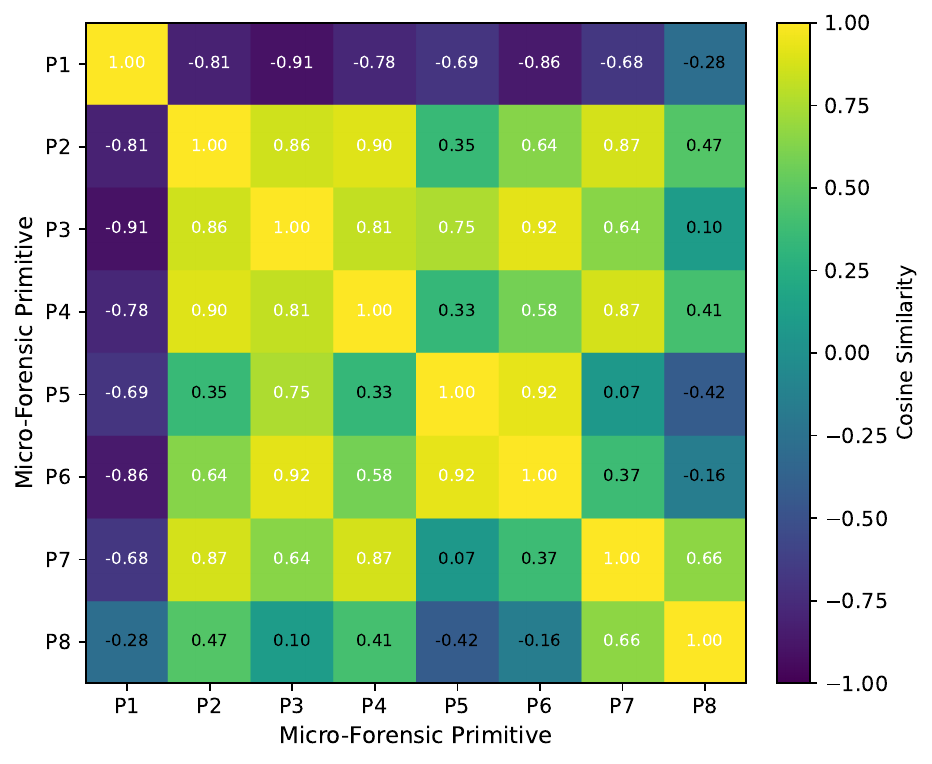}
	\caption{Pairwise cosine similarity matrix of the learned micro-forensic primitives under the $C \;to\; W$ protocol. Most primitive pairs exhibit low similarity, indicating that the learned primitives capture diverse aspects of localized forensic evidence.}
	\label{figprimitive_similarity_matrix}
\end{figure}

To quantify the representational diversity of the learned micro-forensic primitives, we compute the pairwise cosine similarity among the primitives in the final layer under the $C \;to\; W$ protocol, as shown in \figurename~\ref{figprimitive_similarity_matrix}. Most primitive pairs exhibit relatively low similarity, and none shows an excessively high correlation. This suggests that the primitive library avoids evident representational collapse and severe redundancy. Instead, the primitives occupy differentiated directions in the feature space, indicating that joint optimization encourages them to capture distinct aspects of local forensic evidence. Such diversity provides a rich candidate evidence pool for subsequent adaptive routing and compositional inference.

\subsubsection{Primitive Utilization Analysis}

\begin{figure}[tpb]
	\centering
	\includegraphics[width=0.5\textwidth]{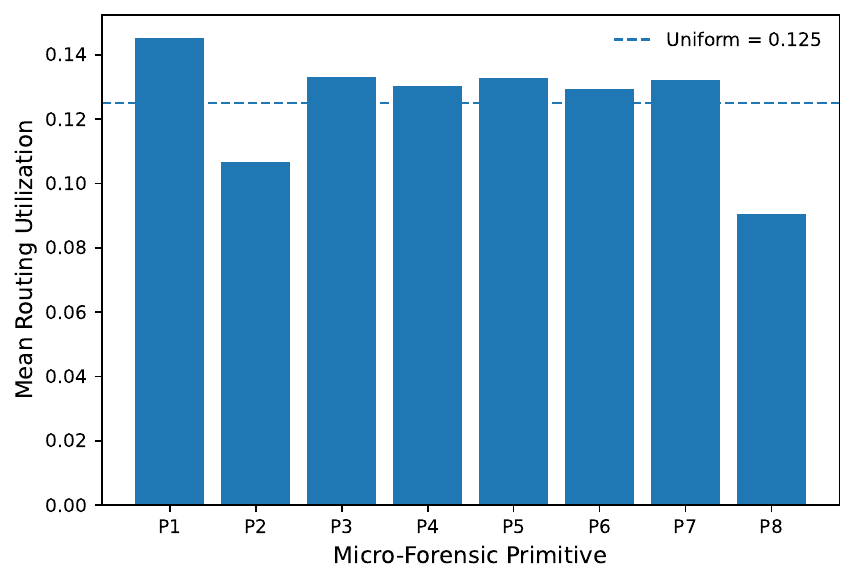}
	\caption{Utilization of the micro-forensic primitives under the $C \;to\; W$ protocol. All primitives remain actively utilized, while their different average routing weights reflect differentiated contributions to compositional forensic representation.}
	\label{figglobal_primitive_utilization}
\end{figure}

To investigate how the learned micro-forensic primitives are actually utilized during test-time compositional inference, we analyze the routing weights of the last layer conditioned on the global contextual prompt associated with the predicted class of each sample under the $C \;to\; W$ protocol. Specifically, for each test sample, we select the routing distribution corresponding to its predicted class and average the routing weights of each primitive over the entire target domain to obtain its utilization. As shown in \figurename~\ref{figglobal_primitive_utilization}, all eight primitives receive non-negligible utilization, with average weights ranging from $0.0906$ to $0.1444$. In particular, $P_1$ exhibits the highest utilization, whereas $P_8$ receives the lowest but remains actively involved in the compositional representation. The remaining primitives show relatively balanced yet differentiated utilization around the uniform reference of $1/8=0.125$. These results indicate that the routing mechanism does not collapse onto a small subset of primitives; instead, the learned primitive library is broadly utilized while different primitives contribute with different strengths to forensic evidence composition.

\subsection{Comparison with State-of-the-Art Methods}

\begin{table}
	\small
	\caption{Comparison with state-of-the-art methods using CASIA-SURF (S) dataset as the source domain and CeFA (F) dataset as the target domain. A lower HTER indicates better performance, while a higher AUC signifies better performance.}
	\label{tabsf}
	\centering
	\rowcolors{2}{white}{blue!5}
	 \begin{tabular}{p{2.5cm}p{2.5cm}<{\centering}p{2.5cm}<{\centering}}
		\hline
		Method& \multicolumn{2}{c}{$S \;to\; F$} \\
		& HTER (\%)&AUC (\%)   \\
		MS-LBP \cite{maatta2011face} &53.80 &46.52\\
		Color texture  \cite{boulkenafet2016face} &55.85 &45.55\\
		CNN	 \cite{yang2014learn} &43.74 &61.07\\
		DTN\cite{liu2019deep}  &41.37 &64.45\\
		SDTN\cite{liu2020disentangling}  & 36.03&66.47\\
		DR-UDA\cite{wang2020unsupervised}  &45.58& 57.34\\
		USDAN\cite{jia2021unified}  &43.04& 58.26\\
		PatchCNN\cite{wang2022patchnet}  &37.92& 63.83\\
		PDEN\cite{li2021progressive}  &55.43 &42.32\\
		LD\cite{wang2021learning}  &44.45 &58.56\\
		IADG\cite{zhou2023instance}&47.70&50.11\\ 
		OSDG\cite{jiang2024open}  &33.91 &71.42\\
		FoundPAD\cite{ozgur2025foundpad} &44.90&58.11\\
		MEFAS\cite{cai2025me}&29.42  & 76.95\\ 
		MVPFAS\cite{yu2025multi}&39.90 & 64.98\\ 
		Ours&\textbf{25.31}&\textbf{79.44}\\
		\hline
		
	\end{tabular}
\end{table}

We compare the proposed method with representative face anti-spoofing approaches on nine open-world protocols, including handcrafted-feature methods, conventional deep models, domain adaptation, domain generalization methods, and foundation-model adaptation approaches. The results are reported in Tables~\ref{tabw}, \ref{tabh}, and \ref{tabsf}. Handcrafted descriptors, such as MS-LBP and Color Texture, and conventional convolutional models exhibit substantial performance degradation when covariate and semantic shifts occur simultaneously. Domain adaptation and generalization methods, including DTN, USDAN, PatchCNN, and OSDG, partially mitigate this degradation by learning more transferable domain-invariant representations. Methods that adapt pretrained foundation models through LoRA or prompt learning, such as FoundPAD, MEFAS, and MVPFAS, generally achieve stronger performance, highlighting the value of transferable priors learned from large-scale pretraining.

The proposed method delivers the best or highly competitive performance across the reported protocols. In Table~\ref{tabw}, it achieves the lowest average HTER of 22.26\% and the highest average AUC of 83.44\%, corresponding to a 29.51\% relative reduction in HTER over the second-best method, MEFAS. It likewise obtains the lowest average HTER and highest average AUC in Tables~\ref{tabh} and \ref{tabsf}. These results demonstrate that the learned compositional forensic visual prompts generalize well to simultaneous covariate and semantic shifts induced by cross-domain heterogeneous unseen attacks.

The proposed method also exhibits stronger cross-protocol consistency. In contrast, several competing methods perform well under specific settings but degrade markedly when the source distribution changes. For example, MEFAS achieves a high AUC on $C \;to\; H$ but performs substantially worse on $M \;to\; H$, suggesting that prompts learned from source datasets with limited attack diversity may not adequately cover the heterogeneous attacks in the target domain. Overall, the consistent gains across protocols indicate that the proposed method better suppresses domain-specific interference and captures transferable attack-discriminative evidence, resulting in improved robustness in open-world scenarios.

\subsection{Ablation Study and Hyperparameter Analysis}
\subsubsection{Component Contribution Analysis}
We evaluate the contribution of each component through the ablations reported in Table~\ref{tabcc}. The variant \textit{w/o global augmentation} removes the augmentation between the global contextual and primitive evidence prompts in Equation~\ref{equ1}, while retaining global guidance for primitive composition. The variant \textit{w/o global guide} directly aggregates all refined primitives without class-specific global routing. The variant \textit{w/o all global contextual prompts} removes the global contextual prompts entirely and performs classification using only the aggregated primitive evidence.

Removing the final global-context augmentation increases HTER by less than one percentage point, indicating that this residual augmentation provides additional improvement but is not the primary source of performance gains. In contrast, removing global guidance increases HTER by approximately seven percentage points. This demonstrates that different samples and attack types require adaptive selection and composition of relevant primitives, and simple aggregation is insufficient to characterize the diverse forensic evidence in heterogeneous attacks.
When all global contextual prompts are removed, the model nearly loses its discriminative capability, suggesting that the primitives require class-specific global priors to form meaningful real and spoof representations.
This also suggests that the primitives are inherently shared and reusable across categories rather than being directly aligned with specific classes. 
Removing the primitive evidence prompts increases HTER by approximately 16 percentage points, confirming that localized primitive evidence is the principal source of discriminative information. Overall, these results reveal a clear division of roles: micro-forensic primitives provide fine-grained and transferable evidence, whereas global contextual prompts supply class-specific physical priors for dynamic routing and composition.

\subsubsection{Effect of the Number of Primitives}
To examine the effect of primitive-library size, we vary the number of micro-forensic primitives from 4 to 12. As shown in Table~\ref{tabnp}, performance first improves and then declines. An insufficient number of primitives limits the diversity of forensic evidence that can be represented, whereas an excessively large primitive set increases composition complexity and introduces redundant or irrelevant responses. We therefore use eight primitives, which provides the best trade-off between evidence coverage and composition stability.

\begin{table}
	\small
	\caption{Component contribution analysis results under the {$C \;to\; W$} protocol.}
	\label{tabcc}
	\centering
	\rowcolors{2}{white}{blue!5}
	\begin{tabular}{p{4.4cm}p{1.4cm}<{\centering}p{1.4cm}<{\centering}}
		\hline
		Component& HTER (\%)&AUC (\%)   \\
		\textit{w/o primitive evidence prompts}&26.72&80.72\\
		\textit{w/o global augmentation}&10.89&95.08\\
		\textit{w/o global guild}&17.84 &90.29\\
		\textit{w/o all global contextual prompts}&50.00&50.00\\
		Ours&\textbf{10.14}&\textbf{95.26}\\
		\hline
	\end{tabular}
\end{table}

\begin{table}[t]
	\small
	\caption{Comparison of the number of primitives under the {$C \;to\; W$} protocol.}
	\label{tabnp}
	\centering
	\rowcolors{2}{white}{blue!5}
	 \begin{tabular}{p{3.5cm}<{\centering}p{2cm}<{\centering}p{2cm}<{\centering}}
		\hline
		Number of Primitives& HTER (\%)&AUC (\%)   \\
		4 &16.78 &90.16\\
		6 &12.51 &93.86\\
		8 & \textbf{10.14}&\textbf{95.26}\\
		10 &12.57 &92.74\\
		12 & 15.56&91.02\\
		\hline
	\end{tabular}
\end{table}

\begin{table}[t]
	\small
	\caption{Ablation analysis of visual prompts at different layers under the {$C \;to\; W$} protocol.}
	\label{tabsfdl}
	\centering
	\rowcolors{2}{white}{blue!5}
	\begin{tabular}{p{3.5cm}<{\centering}p{2cm}<{\centering}p{2cm}<{\centering}}
		\hline
		Component& HTER (\%)&AUC (\%)   \\
		w/o layer6&15.45&92.32\\
		w/o layer12&11.89&92.93\\
		w/o layer18&15.00&91.97\\
		w/o layer24&11.13&94.37\\
		Ours&\textbf{10.14}&\textbf{95.26}\\
		\hline	
	\end{tabular}
\end{table}

\begin{table}[t]
	\small
	\caption{Comparison of vision foundation model backbones under the {$C \;to\; W$} protocol.}
	\label{tabbs}
	\centering
	\rowcolors{2}{white}{blue!5}
	\begin{tabular}{p{3.5cm}<{\centering}p{2cm}<{\centering}p{2cm}<{\centering}}
		\hline
		Backbone& HTER (\%)&AUC (\%)   \\
		ViT-B/16  & 23.40&83.09\\
		ViT-B/32 & 24.69&82.59\\
		ViT-L/14 &13.61 &92.91\\
		ViT-L/14@336px & \textbf{10.14}&\textbf{95.26}\\
		\hline
	\end{tabular}
\end{table}

\subsubsection{Ablation of Layer-Specific Visual Prompts}
We assess the contribution of visual prompts inserted at different stages by removing them individually. As shown in Table~\ref{tabsfdl}, removing the prompts at Layers 6 and 18 causes an approximately five percentage point performance drop, whereas removing those at Layers 12 and 24 results in a smaller degradation of about two percentage points. This indicates that the prompts at Layers 6 and 18 contribute more substantially to the final compositional representation.
Prompts at relatively shallow layers are more effective at capturing local textures, boundaries, and high-frequency artifacts, whereas prompts at intermediate and deeper layers emphasize structural and semantic-level cues. Although the prompts at Layers 12 and 24 provide complementary information, part of their evidence may overlap with that captured at other depths. These results confirm that multi-stage prompting integrates complementary forensic evidence across the feature hierarchy.

\subsubsection{Effect of Different Vision Foundation Model Backbones}

We evaluate the impact of different vision foundation model backbones under the $C \;to\; W$ protocol. When using ViT-B as the backbone, the four selected layers are set to 3, 6, 9, and 12. As shown in Table~\ref{tabbs}, the proposed method consistently benefits from stronger visual representations. ViT-B/16 slightly outperforms ViT-B/32, indicating that finer patch granularity facilitates the capture of localized spoofing artifacts. Replacing the base-scale backbone with ViT-L/14 substantially reduces HTER from 23.40\% to 13.61\% and improves AUC from 83.09\% to 92.91\%, highlighting the advantage of increased model capacity for transferable forensic representation learning. ViT-L/14@336px further improves performance, achieving the lowest HTER of 10.14\% and the highest AUC of 95.26\%. These results show that the proposed compositional visual prompting framework can effectively leverage stronger and higher-resolution vision backbones to capture fine-grained and spatially localized forensic evidence.

\subsubsection{Category-Wise Error Rate Analysis}

\begin{figure}[tpb]
	\centering
	\includegraphics[width=0.5\textwidth]{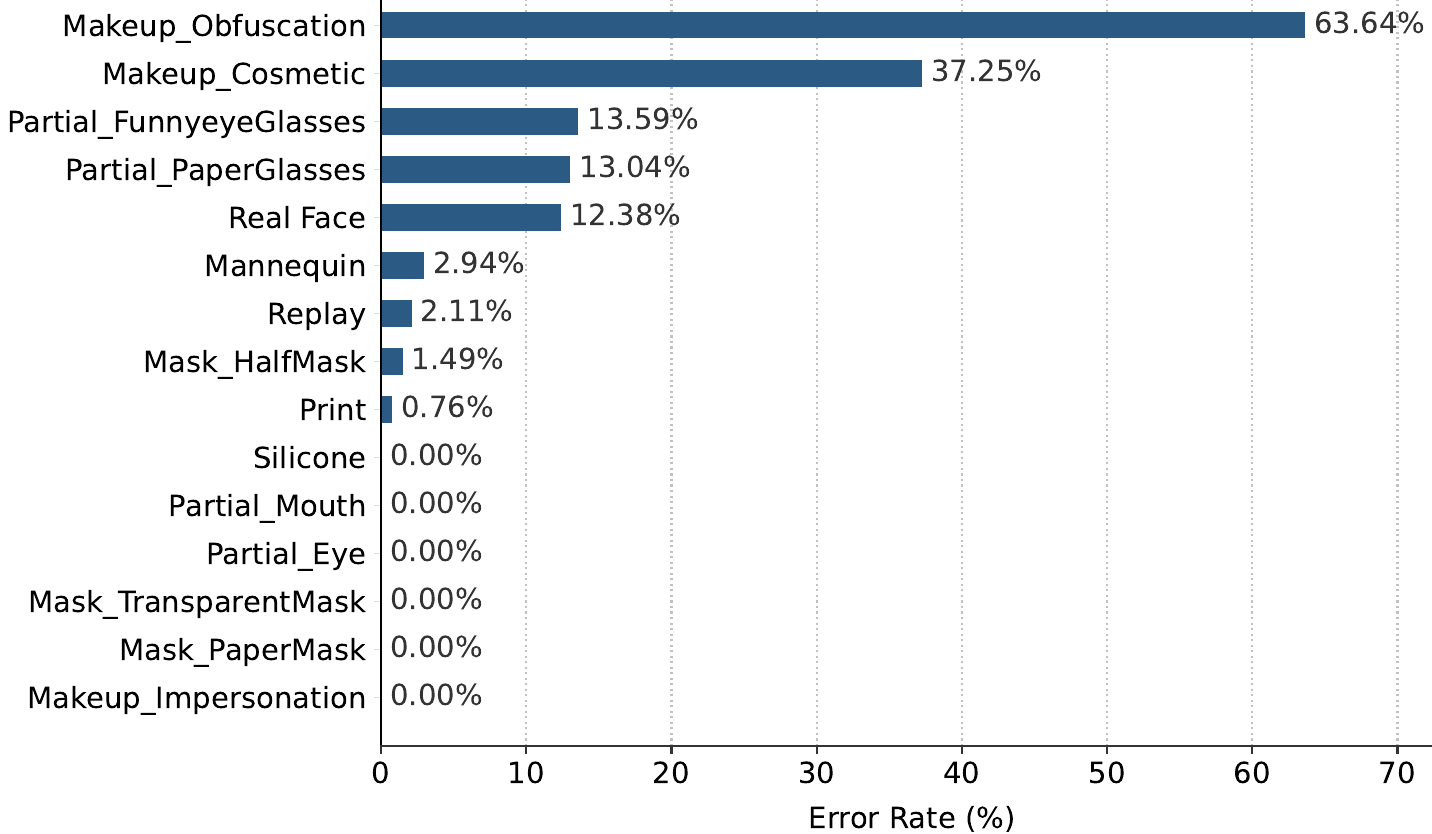}
	\caption{Category-wise error rates under the $C \;to\; W$ protocol. Errors are mainly concentrated in makeup and partial attacks with subtle or spatially localized spoofing cues, while most other attack types achieve error rates below 3\%. 
	}
	\label{figpercat_error_rates}
\end{figure}

We analyze the category-wise error rates under the $C \;to\; W$ protocol to explore how well our method generalizes to different types of attacks, as shown in \figurename~\ref{figpercat_error_rates}. Errors are concentrated in a few challenging categories. In particular, makeup obfuscation and makeup cosmetic exhibit the highest error rates, while partial paper glasses, real faces, and partial funny-eye glasses show moderate errors. These categories largely preserve natural facial appearance and introduce only subtle or spatially localized artifacts, making their forensic cues more difficult to distinguish. In contrast, most remaining attack types achieve error rates below 3\%, as they exhibit more pronounced discrepancies in texture, geometry, reflectance, or presentation medium. These results demonstrate that the learned compositional forensic visual prompts provide robust discrimination across diverse presentation attacks.

\section{Conclusion}

This paper addresses the concurrent covariate and semantic shifts in open-world face anti-spoofing through a compositional forensic visual prompt learning framework. Rather than representing heterogeneous attacks with closed-set textual categories, the proposed method learns reusable forensic primitives directly in the visual feature space and dynamically routes and composes them under global-prior guidance to construct input-adaptive compositional forensic visual prompts.
Extensive experiments on nine open-world protocols demonstrate that the proposed method achieves state-of-the-art performance, with strong cross-domain generalization and adaptability to unseen attacks. Ablation studies and analyses in the frequency, spatial, and feature domains further show that micro-forensic primitives capture transferable and composable local evidence, while global contextual prompts provide class-specific physical priors for input-adaptive primitive composition. The resulting compositional forensic visual prompts enable unified representation and discrimination of heterogeneous seen and unseen attacks. Future work will investigate more explicit semantic interpretations of the learned primitives and extend the framework to multimodal face anti-spoofing.

\bibliographystyle{IEEEtran}
\bibliography{ref2}

@inproceedings{maatta2011face,
  title={Face spoofing detection from single images using micro-texture analysis},
  author={M{\"a}{\"a}tt{\"a}, Jukka and Hadid, Abdenour and Pietik{\"a}inen, Matti},
  booktitle={IEEE International Joint Conference on Biometrics},
  pages={1--7},
  year={2011}
}

@inproceedings{chingovska2012effectiveness,
  title={On the effectiveness of local binary patterns in face anti-spoofing},
  author={Chingovska, Ivana and Anjos, Andr{\'e} and Marcel, S{\'e}bastien},
  booktitle={International Conference of Biometrics Special Interest Group},
  pages={1--7},
  year={2012}
}

@article{boulkenafet2016face,
  title={Face spoofing detection using colour texture analysis},
  author={Boulkenafet, Zinelabidine and Komulainen, Jukka and Hadid, Abdenour},
  journal={IEEE Transactions on Information Forensics and Security},
  volume={11},
  number={8},
  pages={1818--1830},
  year={2016}
}

@inproceedings{zhang2012face,
  title={A face antispoofing database with diverse attacks},
  author={Zhang, Zhiwei and Yan, Junjie and Liu, Sifei and Lei, Zhen and Yi, Dong and Li, Stan Z},
  booktitle={International Conference on Biometrics},
  pages={26--31},
  year={2012}
}

@article{wen2015face,
  title={Face spoof detection with image distortion analysis},
  author={Wen, Di and Han, Hu and Jain, Anil K},
  journal={IEEE Transactions on Information Forensics and Security},
  volume={10},
  number={4},
  pages={746--761},
  year={2015}
}

@article{yang2014learn,
  title={Learn convolutional neural network for face anti-spoofing},
  author={Yang, Jianwei and Lei, Zhen and Li, Stan Z},
  journal={arXiv preprint arXiv:1408.5601},
  year={2014}
}

@inproceedings{boulkenafet2017oulu,
  title={OULU-NPU: A mobile face presentation attack database with real-world variations},
  author={Boulkenafet, Zinelabinde and Komulainen, Jukka and Li, Lei and Feng, Xiaoyi and Hadid, Abdenour},
  booktitle={IEEE International Conference on Automatic Face and Gesture Recognition},
  pages={612--618},
  year={2017}
}

@article{Zhang2016Joint,
  title={Joint face detection and alignment using multitask cascaded convolutional networks},
  author={Zhang, Kaipeng and Zhang, Zhanpeng and Li, Zhifeng and Qiao, Yu},
  journal={IEEE Signal Processing Letters},
  volume={23},
  number={10},
  pages={1499-1503},
  year={2016},
}

@inproceedings{jia2020single,
  title={Single-side domain generalization for face anti-spoofing},
  author={Jia, Yunpei and Zhang, Jie and Shan, Shiguang and Chen, Xilin},
  booktitle={IEEE Conference on Computer Vision and Pattern Recognition},
  pages={8484--8493},
  year={2020}
}

@inproceedings{liu2019deep,
  title={Deep tree learning for zero-shot face anti-spoofing},
  author={Liu, Yaojie and Stehouwer, Joel and Jourabloo, Amin and Liu, Xiaoming},
  booktitle={IEEE Conference on Computer Vision and Pattern Recognition},
  pages={4680--4689},
  year={2019}
}

@inproceedings{liu2021casia,
	title={Casia-surf cefa: A benchmark for multi-modal cross-ethnicity face anti-spoofing},
	author={Liu, Ajian and Tan, Zichang and Wan, Jun and Escalera, Sergio and Guo, Guodong and Li, Stan Z},
	booktitle={IEEE Winter Conference on Applications of Computer Vision},
	pages={1179--1187},
	year={2021}
}

@inproceedings{liu2020disentangling,
  title={On disentangling spoof trace for generic face anti-spoofing},
  author={Liu, Yaojie and Stehouwer, Joel and Liu, Xiaoming},
  booktitle={European Conference on Computer Vision},
  pages={406--422},
  year={2020}
}

@article{zhang2020casia,
  title={Casia-surf: A large-scale multi-modal benchmark for face anti-spoofing},
  author={Zhang, Shifeng and Liu, Ajian and Wan, Jun and Liang, Yanyan and Guo, Guodong and Escalera, Sergio and Escalante, Hugo Jair and Li, Stan Z},
  journal={IEEE Transactions on Biometrics, Behavior, and Identity Science},
  volume={2},
  number={2},
  pages={182--193},
  year={2020},
  publisher={IEEE}
}

@article{yu2021deep,
  title={Deep learning for face anti-spoofing: A survey},
  author={Yu, Zitong and Qin, Yunxiao and Li, Xiaobai and Zhao, Chenxu and Lei, Zhen and Zhao, Guoying},
  journal={ IEEE Transactions on Pattern Analysis and Machine Intelligence},
  year={2023},
  volume={45},
  number={05},
  pages={5609-5631},
}

@article{jia2021dual,
	title={Dual-Branch Meta-Learning Network With Distribution Alignment for Face Anti-Spoofing},
	author={Jia, Yunpei and Zhang, Jie and Shan, Shiguang},
	journal={IEEE Transactions on Information Forensics and Security},
	volume={17},
	pages={138--151},
	year={2021}
}

@article{wang2020unsupervised,
	title={Unsupervised adversarial domain adaptation for cross-domain face presentation attack detection},
	author={Wang, Guoqing and Han, Hu and Shan, Shiguang and Chen, Xilin},
	journal={IEEE Transactions on Information Forensics and Security},
	volume={16},
	pages={56--69},
	year={2020}
}

@article{jia2021unified,
	title={Unified unsupervised and semi-supervised domain adaptation network for cross-scenario face anti-spoofing},
	author={Jia, Yunpei and Zhang, Jie and Shan, Shiguang and Chen, Xilin},
	journal={Pattern Recognition},
	volume={115},
	pages={107888},
	year={2021}
}

@inproceedings{srivatsan2023flip,
	title={FLIP: Cross-domain Face Anti-spoofing with Language Guidance},
	author={Srivatsan, Koushik and Naseer, Muzammal and Nandakumar, Karthik},
	booktitle={International Conference on Computer Vision},
	pages={19685--19696},
	year={2023}
}

@article{Zhou_2022,
	year = 2022,
	volume = {130},
	number = {9},
	pages = {2337--2348},
	author = {Kaiyang Zhou and Jingkang Yang and Chen Change Loy and Ziwei Liu},
	title = {Learning to Prompt for Vision-Language Models},	
	journal = {International Journal of Computer Vision}
}

@inproceedings{narayan2024hyp,
	title={Hyp-oc: Hyperbolic one class classification for face anti-spoofing},
	author={Narayan, Kartik and Patel, Vishal M},
	booktitle={IEEE International Conference on Automatic Face and Gesture Recognition},
	pages={1--10},
	year={2024}
}

@inproceedings{huang2024one,
	title={One-Class Face Anti-spoofing via Spoof Cue Map-Guided Feature Learning},
	author={Huang, Pei-Kai and Chiang, Cheng-Hsuan and Chen, Tzu-Hsien and Chong, Jun-Xiong and Liu, Tyng-Luh and Hsu, Chiou-Ting},
	booktitle={IEEE Conference on Computer Vision and Pattern Recognition},
	pages={277--286},
	year={2024}
}

@article{liu2025dual,
	title={Dual Consistency Regularization for Generalized Face Anti-Spoofing},
	author={Liu, Yongluo and Li, Zun and Wu, Lifang},
	journal={IEEE Transactions on Information Forensics and Security},
	year={2025}
}

@inproceedings{ge2024difffas,
	title={Difffas: face anti-spoofing via generative diffusion models},
	author={Ge, Xinxu and Liu, Xin and Yu, Zitong and Shi, Jingang and Qi, Chun and Li, Jie and K{\"a}lvi{\"a}inen, Heikki},
	booktitle={European Conference on Computer Vision},
	pages={144--161},
	year={2024}
}

@inproceedings{guo2022multi,
	title={Multi-domain learning for updating face anti-spoofing models},
	author={Guo, Xiao and Liu, Yaojie and Jain, Anil and Liu, Xiaoming},
	booktitle={European Conference on Computer Vision},
	pages={230--249},
	year={2022}
}

@article{jiang2023adversarial,
	title={Adversarial learning domain-invariant conditional features for robust face anti-spoofing},
	author={Jiang, Fangling and Li, Qi and Liu, Pengcheng and Zhou, Xiang-Dong and Sun, Zhenan},
	journal={International Journal of Computer Vision},
	volume={131},
	pages={1680--1703},
	year={2023}
}

@inproceedings{hu2024fine,
	title={Fine-grained prompt learning for face anti-spoofing},
	author={Hu, Xueli and Liu, Huan and Yuan, Haocheng and Fu, Zhiyang and Luo, Yizhi and Zhang, Ning and Zou, Hang and Gan, Jianwen and Zhang, Yuan},
	booktitle={ACM International Conference on Multimedia},
	pages={7619--7628},
	year={2024}
}

@inproceedings{guo2024style,
	title={Style-conditional prompt token learning for generalizable face anti-spoofing},
	author={Guo, Jiabao and Liu, Huan and Luo, Yizhi and Hu, Xueli and Zou, Hang and Zhang, Yuan and Liu, Hui and Zhao, Bo},
	booktitle={ACM International Conference on Multimedia},
	pages={994--1003},
	year={2024}
}

@inproceedings{wang2024tf,
	title={TF-FAS: twofold-element fine-grained semantic guidance for generalizable face anti-spoofing},
	author={Wang, Xudong and Zhang, Ke-Yue and Yao, Taiping and Zhou, Qianyu and Ding, Shouhong and Dai, Pingyang and Ji, Rongrong},
	booktitle={European Conference on Computer Vision},
	pages={148--168},
	year={2024}
}

@inproceedings{wang2025faceshield,
	title={Faceshield: Explainable face anti-spoofing with multimodal large language models},
	author={Wang, Hongyang and Shi, Yichen and Tao, Zhuofu and Gao, Yuhao and Zhang, Liepiao and Lin, Xun and Feng, Jun and Yuan, Xiaochen and Yu, Zitong and Cao, Xiaochun},
	booktitle={AAAI Conference on Artificial Intelligence},
	volume={40},
	number={12},
	pages={9811--9819},
	year={2026}
}

@inproceedings{liu2024bottom,
	title={Bottom-up domain prompt tuning for generalized face anti-spoofing},
	author={Liu, Si-Qi and Wang, Qirui and Yuen, Pong C},
	booktitle={European Conference on Computer Vision},
	pages={170--187},
	year={2024}
}

@article{kong2024dual,
	title={Dual teacher knowledge distillation with domain alignment for face anti-spoofing},
	author={Kong, Zhe and Zhang, Wentian and Wang, Tao and Zhang, Kaihao and Li, Yuexiang and Tang, Xiaoying and Luo, Wenhan},
	journal={IEEE Transactions on Circuits and Systems for Video Technology},
	year={2024}
}

@article{cai2024towards,
	title={Towards Data-Centric Face Anti-spoofing: Improving Cross-Domain Generalization via Physics-Based Data Synthesis},
	author={Cai, Rizhao and Soh, Cecelia and Yu, Zitong and Li, Haoliang and Yang, Wenhan and Kot, Alex C},
	journal={International Journal of Computer Vision},
	pages={1--22},
	year={2024}
}

@inproceedings{le2024gradient,
	title={Gradient alignment for cross-domain face anti-spoofing},
	author={Le, Binh M and Woo, Simon S},
	booktitle={IEEE Conference on Computer Vision and Pattern Recognition},
	pages={188--199},
	year={2024}
}

@inproceedings{yang2024generalized,
	title={Generalized Face Anti-spoofing via Finer Domain Partition and Disentangling Liveness-irrelevant Factors},
	author={Yang, Jingyi and Yu, Zitong and Ni, Xiuming and He, Jia and Li, Hui},
	booktitle={European Conference on Artificial Intelligence},
	pages={274--281},
	year={2024}
}

@article{ma2024dual,
	title={Dual feature disentanglement for face anti-spoofing},
	author={Ma, Yimei and Qian, Jianjun and Li, Jun and Yang, Jian},
	journal={Pattern Recognition},
	volume={155},
	pages={110656},
	year={2024},
	publisher={Elsevier}
}

@inproceedings{ozgur2025foundpad,
	title={FoundPAD: Foundation models reloaded for face presentation attack detection},
	author={Ozgur, Guray and Caldeira, Eduarda and Chettaoui, Tahar and Boutros, Fadi and Ramachandra, Raghavendra and Damer, Naser},
	booktitle={Winter Conference on Applications of Computer Vision},
	pages={745--755},
	year={2025}
}

@inproceedings{li2025optimal,
	title={Optimal Transport-Guided Source-Free Adaptation for Face Anti-Spoofing},
	author={Li, Zhuowei and Zhao, Tianchen and Xu, Xiang and Zhang, Zheng and Li, Zhihua and Chen, Xuanbai and Zhang, Qin and Bergamo, Alessandro and Jain, Anil K and Xing, Yifan},
	booktitle={Computer Vision and Pattern Recognition Conference},
	pages={24351--24363},
	year={2025}
}

@inproceedings{huang2023test,
	title={Test-Time Adaptation for Robust Face Anti-Spoofing.},
	author={Huang, Pei-Kai and Lu, Chen-Yu and Chang, Shu-Jung and Chong, Jun-Xiong and Hsu, Chiou-Ting},
	booktitle={British Machine Vision Conference},
	pages={379--380},
	year={2023}
}

@article{jiang2024open,
	title={Open-Set Single-Domain Generalization for Robust Face Anti-Spoofing},
	author={Jiang, Fangling and Li, Qi and Wang, Weining and Ren, Min and Shen, Wei and Liu, Bing and Sun, Zhenan},
	journal={International Journal of Computer Vision},
	volume={132},
	number={11},
	pages={5151--5172},
	year={2024}
}

@inproceedings{huang2025slip,
	title={SLIP: Spoof-aware one-class face anti-spoofing with language image pretraining},
	author={Huang, Pei-Kai and Chong, Jun-Xiong and Chiang, Cheng-Hsuan and Chen, Tzu-Hsien and Liu, Tyng-Luh and Hsu, Chiou-Ting},
	booktitle={Association for the Advancement of Artificial Intelligence},
	volume={39},
	pages={3697--3706},
	year={2025}
}

@inproceedings{dong2021open,
	title={Open set face anti-spoofing in unseen attacks},
	author={Dong, Xin and Liu, Hao and Cai, Weiwei and Lv, Pengyuan and Yu, Zekuan},
	booktitle={ACM International Conference on Multimedia},
	pages={4082--4090},
	year={2021}
}

@inproceedings{yu2025multi,
	title={Multi-View Slot Attention Using Paraphrased Texts for Face Anti-Spoofing},
	author={Yu, Jeongmin and Kim, Susang and Lee, Kisu and Kwon, Taekyoung and Shin, Won-Yong and Kim, Ha Young},
	booktitle={International Conference on Computer Vision},
	pages={21117--21128},
	year={2025}
}

@article{cai2025me,
	title={ME-FAS: Multimodal Text Enhancement for Cross-Domain Face Anti-Spoofing},
	author={Cai, Lvpan and Wang, Haowei and Ji, Jiayi and Sun, Xiaoshuai and Cao, Liujuan and Ji, Rongrong},
	journal={IEEE Transactions on Information Forensics and Security},
	year={2025},
	publisher={IEEE}
}

@inproceedings{zhou2023instance,
	title={Instance-aware domain generalization for face anti-spoofing},
	author={Zhou, Qianyu and Zhang, Ke-Yue and Yao, Taiping and Lu, Xuequan and Yi, Ran and Ding, Shouhong and Ma, Lizhuang},
	booktitle={Conference on Computer Vision and Pattern Recognition},
	pages={20453--20463},
	year={2023}
}

@inproceedings{zhang2025interpretable,
	title={Interpretable face anti-spoofing: Enhancing generalization with multimodal large language models},
	author={Zhang, Guosheng and Wang, Keyao and Yue, Haixiao and Liu, Ajian and Zhang, Gang and Yao, Kun and Ding, Errui and Wang, Jingdong},
	booktitle={AAAI Conference on Artificial Intelligence},
	volume={39},
	number={9},
	pages={9896--9904},
	year={2025}
}

@InProceedings{wang2022patchnet,
	author    = {Wang, Chien-Yi and Lu, Yu-Ding and Yang, Shang-Ta and Lai, Shang-Hong},
	booktitle = {Conference on Computer Vision and Pattern Recognition},
	title     = {PatchNet: A simple face anti-spoofing framework via fine-grained patch recognition},
	pages     = {20281--20290},
	year      = {2022}
}

@inproceedings{li2021progressive,
	title={Progressive domain expansion network for single domain generalization},
	author={Li, Lei and Gao, Ke and Cao, Juan and Huang, Ziyao and Weng, Yepeng and Mi, Xiaoyue and Yu, Zhengze and Li, Xiaoya and Xia, Boyang},
	booktitle={Conference on Computer Vision and Pattern Recognition},
	pages={224--233},
	year={2021}
}

@inproceedings{wang2021learning,
	title={Learning to diversify for single domain generalization},
	author={Wang, Zijian and Luo, Yadan and Qiu, Ruihong and Huang, Zi and Baktashmotlagh, Mahsa},
	booktitle={International Conference on Computer Vision},
	pages={834--843},
	year={2021}
}

@Article{9146362hqwmca,
	author  = {Heusch, Guillaume and George, Anjith and Geissbühler, David and Mostaani, Zohreh and Marcel, Sébastien},
	title   = {Deep models and shortwave infrared information to detect face presentation attacks},
	number  = {4},
	pages   = {399-409},
	volume  = {2},
	journal = {IEEE Transactions on Biometrics, Behavior, and Identity Science},
	year    = {2020}
}

@article{yang2026fine,
	title={Fine-Grained Domain Alignment for Face Anti-Spoofing with Asymmetric Pseudo-Labels},
	author={Yang, Jing and Cui, Xusheng and Chen, Yuehai and Du, Shaoyi and Chen, Badong and Liu, Yuewen},
	journal={IEEE Transactions on Information Forensics and Security},
	year={2026},
	publisher={IEEE}
}

@article{mao2025weighted,
	title={Weighted joint distribution optimal transport based domain adaptation for cross-scenario face anti-spoofing},
	author={Mao, Shiyun and Chen, Ruolin and Li, Huibin},
	journal={International Journal of Computer Vision},
	volume={133},
	number={2},
	pages={590--610},
	year={2025}
}

@inproceedings{jung2025group,
	title={Group-wise Scaling and Orthogonal Decomposition for Domain-Invariant Feature Extraction in Face Anti-Spoofing},
	author={Jung, Seungjin and Lee, Kanghee and Jeong, Yonghyun and Noh, Haeun and Lee, Jungmin and Choi, Jongwon},
	booktitle={International Conference on Computer Vision},
	pages={13372--13381},
	year={2025}
}

@article{lin2025reliable,
	title={Reliable and balanced transfer learning for generalized multimodal face anti-spoofing},
	author={Lin, Xun and Liu, Ajian and Yu, Zitong and Cai, Rizhao and Wang, Shuai and Yu, Yi and Wan, Jun and Lei, Zhen and Cao, Xiaochun and Kot, Alex},
	journal={IEEE Transactions on Pattern Analysis and Machine Intelligence},
	year={2025}
}

@article{cai2025rehearsal,
	title={Rehearsal-free and efficient continual learning for cross-domain face anti-spoofing},
	author={Cai, Rizhao and Cui, Yawen and Yu, Zitong and Lin, Xun and Chen, Changsheng and Kot, Alex},
	journal={IEEE Transactions on Pattern Analysis and Machine Intelligence},
	year={2025}
}

@inproceedings{cao2025towards,
	title={Towards generalized face anti-spoofing from a frequency shortcut view},
	author={Cao, Junyi and Ma, Chao},
	booktitle={Winter Conference on Applications of Computer Vision},
	pages={1005--1015},
	year={2025}
}

@article{ma2026purify,
	title={Purify then Guide: Rethinking Domain Generalization for Multimodal Face Anti-Spoofing},
	author={Ma, Yingjie and Lin, Xun and Yu, Zitong and Wang, Haonan and Zhang, Ruixin and Ding, Shouhong and Liu, Xin and Yuan, Xiaochen and Xie, Weicheng and Shen, Linlin},
	journal={arXiv preprint arXiv:2505.09484},
	year={2026}
}

@inproceedings{chen2025crossgap,
	title={CrossGAP: Unified Face Anti-Spoofing via Cross-Modal Global-Aware Prompting},
	author={Chen, Xihong and Xie, Yutong and Ma, Hui and Li, Ning and Liang, Yanyan and Guo, Jiabao},
	booktitle={International Conference on Computer Vision},
	pages={3208--3215},
	year={2025}
}

@article{jiang2026learning,
	title={Learning Unknown Spoof Prompts for Generalized Face Anti-Spoofing Using Only Real Face Images: F. Jiang et al.},
	author={Jiang, Fangling and Li, Qi and Wang, Weining and Shen, Wei and Liu, Bing and Sun, Zhenan},
	journal={International Journal of Computer Vision},
	volume={134},
	number={5},
	pages={250},
	year={2026}
}

@inproceedings{zhang2026intuition,
	title={From Intuition to Investigation: A Tool-Augmented Reasoning MLLM Framework for Generalizable Face Anti-Spoofing},
	author={Zhang, Haoyuan and Wang, Keyao and Zhang, Guosheng and Yue, Haixiao and Tan, Zhiwen and Peng, Siran and Zhang, Tianshuo and Tan, Xiao and Chen, Kunbin and He, Wei and others},
	booktitle={Conference on Computer Vision and Pattern Recognition},
	pages={40855--40865},
	year={2026}
}

@inproceedings{zhang2026harnessing,
	title={Harnessing Chain-of-Thought Reasoning in Multimodal Large Language Models for Face Anti-Spoofing},
	author={Zhang, Honglu and Fang, Zhiqin and Zhao, Ningning and Hou, Saihui and Ma, Long and Pei, Renwang and He, Zhaofeng},
	booktitle={Conference on Computer Vision and Pattern Recognition},
	pages={33566--33576},
	year={2026}
}

@inproceedings{ma2026pa,
	title={PA-FAS: Towards Interpretable and Generalizable Multimodal Face Anti-Spoofing via Path-Augmented Reinforcement Learning},
	author={Ma, Yingjie and Lin, Xun and Xu, Yong and Xie, Weicheng and Yu, Zitong},
	booktitle={AAAI Conference on Artificial Intelligence},
	volume={40},
	number={10},
	pages={7856--7864},
	year={2026}
}

@article{zhang2026long,
	title={Long-FAS: Cross-domain face anti-spoofing with long text guidance},
	author={Zhang, Jianwen and Zhang, Jianfeng and Yang, Dedong and Li, Rongtao and Li, Ziyang},
	journal={Image and Vision Computing},
	pages={105901},
	year={2026}
}

@article{guo2025domain,
	title={Domain generalization for face anti-spoofing via content-aware composite prompt engineering},
	author={Guo, Jiabao and Liu, Ajian and Diao, Yunfeng and Zhang, Jin and Ma, Hui and Zhao, Bo and Hong, Richang and Wang, Meng},
	journal={IEEE Transactions on Multimedia},
	year={2025}
}

@article{liu2026dgpdl,
	title={DGPDL: Domain-Guided Prompt Distribution Learning for Generalizable Face Anti-spoofing},
	author={Liu, Ajian and Lin, Xun and Zhi, Ruicong and Liang, Yanyan and Zhu, Xinshan and Cai, Zhanchuan and Wan, Jun and Escalera, Sergio and Lei, Zhen},
	journal={IEEE Transactions on Pattern Analysis and Machine Intelligence},
	year={2026}
}

@article{liu2026icpe,
	title={ICPE-FAS: Instance and Category Prompts Engineering for Generalizable Face Anti-Spoofing: A. Liu et al.},
	author={Liu, Ajian and Lin, Xun and Ma, Hui and Yu, Xinxing and Guo, Jiabao and Yu, Zitong and Wan, Jun and Cai, Zhanchuan and Lei, Zhen and Liang, Yanyan},
	journal={International Journal of Computer Vision},
	volume={134},
	number={6},
	pages={292},
	year={2026}
}

@inproceedings{ko2025text,
	title={Text-to-image synthesis for domain generalization in face anti-spoofing},
	author={Ko, Naeun and Jeong, Yonghyun and Ye, Jong Chul},
	booktitle={Winter Conference on Applications of Computer Vision},
	pages={1850--1860},
	year={2025}
}

@article{zhang2026cpg,
	title={CPG-PAD: Concept-Informed Prompts Guided Presentation Attack Detection},
	author={Zhang, Haoyuan and Zhu, Xiangyu and Gao, Li and Liu, Ajian and Peng, Siran and Lei, Zhen},
	journal={arXiv preprint arXiv:2607.01303},
	year={2026}
}

@article{li2025fine,
	title={Fine-Grained Textual Guidance for Generalized Multi-Modal Face Anti-Spoofing},
	author={Li, Daiyuan and Yu, Zitong and Hu, Jinwu and Chen, Guohao and Zeng, Jinghui and Tan, Mingkui},
	journal={IEEE Transactions on Information Forensics and Security},
	year={2025}
}

@article{ma2026clip,
	title={CLIP-SA: CLIP-Guided Semantic Alignment for Generalizable Face Anti-spoofing},
	author={Ma, Hui and Liu, Ajian and Xue, Shuai and Diao, Yunfeng and Liang, Dashuang and Liang, Hailun and Liang, Yanyan and Wan, Jun and Yang, Xun and Hu, ZhenZhen and others},
	journal={IEEE Transactions on Multimedia},
	year={2026},
	publisher={IEEE}
}

@ARTICLE{11194267,
	author={Jiang, Fangling and Li, Qi and Liu, Bing and Wang, Weining and Shan, Caifeng and Sun, Zhenan and Yang, Ming-Hsuan},
	journal={IEEE Transactions on Pattern Analysis and Machine Intelligence}, 
	title={Learning Knowledge-Based Prompts for Robust 3D Mask Presentation Attack Detection}, 
	year={2026},
	volume={48},
	number={2},
	pages={1321-1338}
}


\vspace{-12mm}
\begin{IEEEbiography}
[{\includegraphics[width=1in,height=1.25in,clip,keepaspectratio]{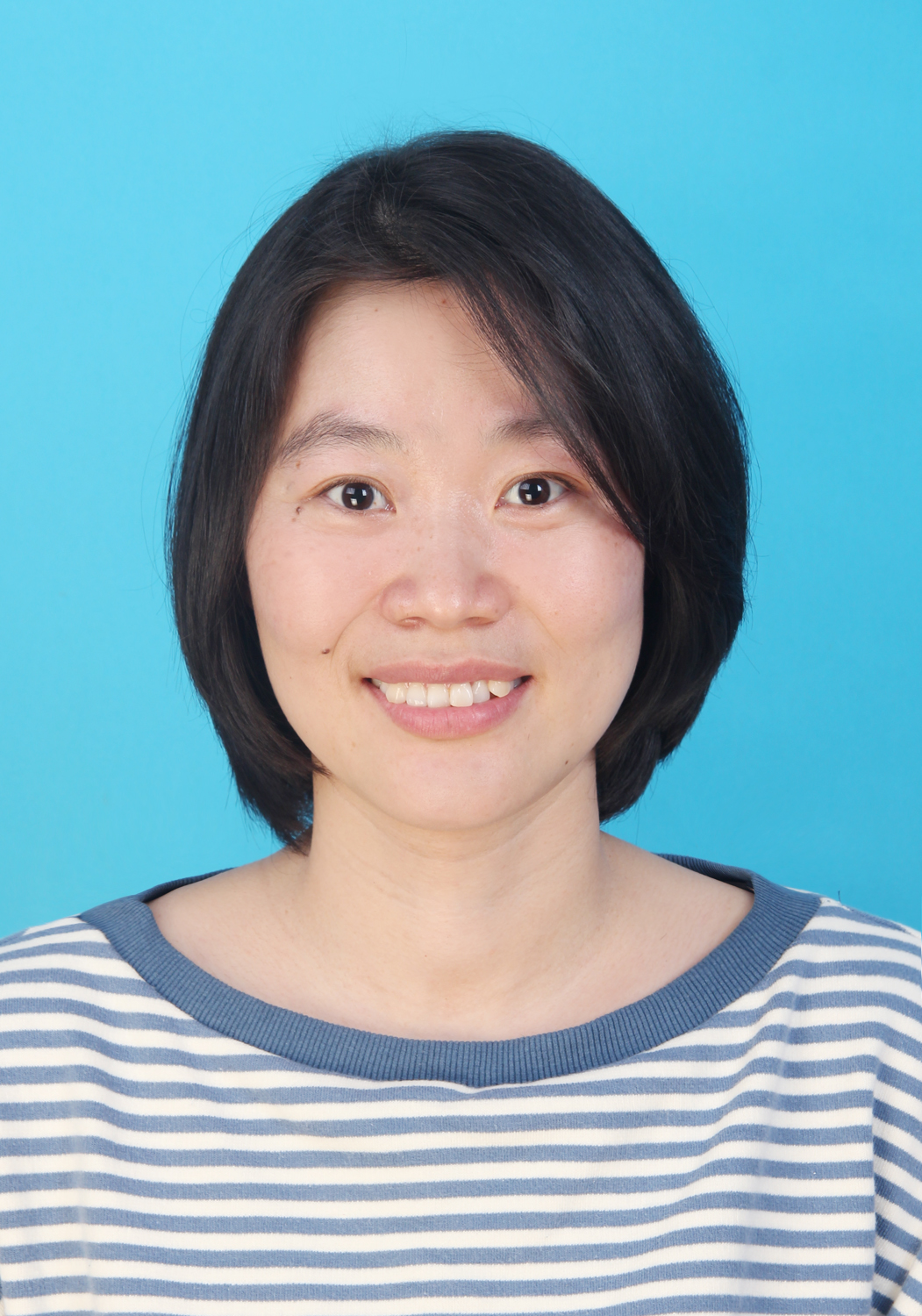}}]
{Fangling Jiang} received the B.E. and M.S. degrees from Tianjin University in 2009 and 2012, respectively, and the Ph.D. degree from University of Chinese Academy of Sciences in 2021. She is an Associate Professor with The First Affiliated Hospital and the School of Computer Science, University of South China. Her research interests include face anti-spoofing, transfer learning, and computer vision.
\end{IEEEbiography}

\vspace{-12mm}
\begin{IEEEbiography}
[{\includegraphics[width=1in,height=1.25in,clip,keepaspectratio]{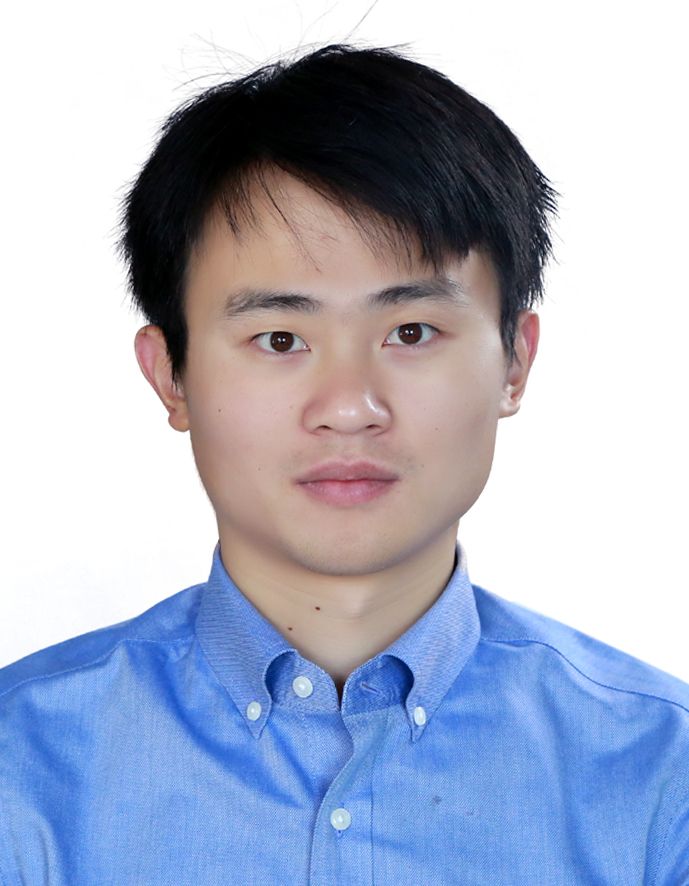}}]
{Qi Li} received the B.E. degree from China University of Petroleum in 2011, the Ph.D. degree from the Institute of Automation, Chinese Academy of Sciences (CASIA) in 2016. 
He is an Associate Professor with the New Laboratory of Pattern Recognition (NLPR), State Key Laboratory of Multimodal Artificial Intelligence Systems (MAIS), CASIA. 
  His research interests include face recognition, computer vision, and machine learning.
\end{IEEEbiography}

\vspace{-12mm}
\begin{IEEEbiography}
[{\includegraphics[width=1in,height=1.25in,clip,keepaspectratio]{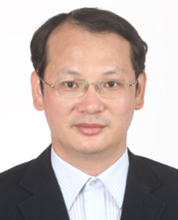}}]
{Bing Liu} is a Professor of Computer Science and Technology at University of South China. Liu serves as a project head of Industry Academia Collaborative Education of the Ministry of Education in 2024. He is the executive director of the Hunan Provincial University Network Association. His research interests include computer vision and machine learning.
\end{IEEEbiography}

\vspace{-12mm}
\begin{IEEEbiography}[{\includegraphics[width=1in,height=1.25in,clip,keepaspectratio]{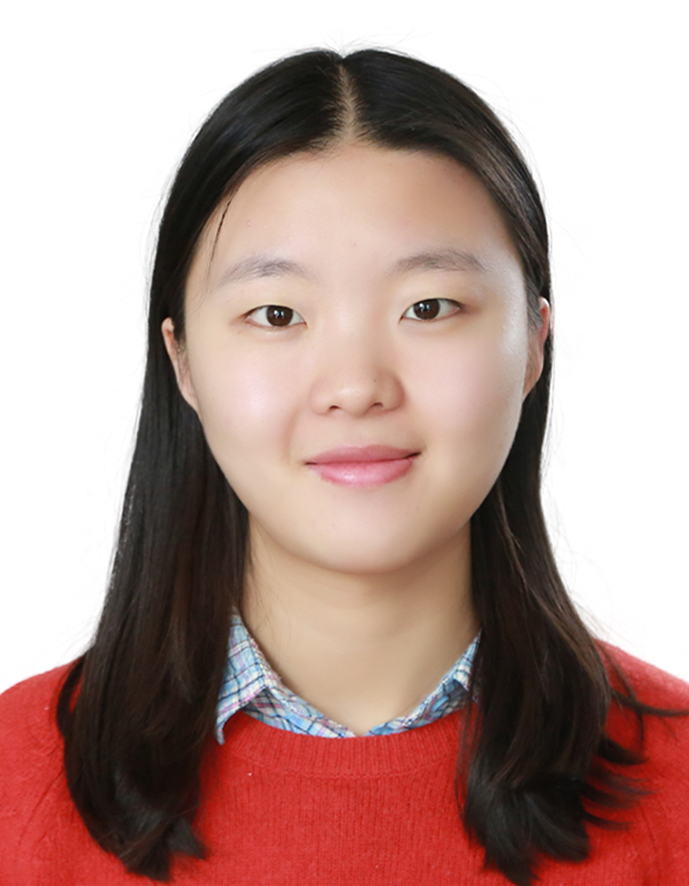}}]{Weining Wang}
  received her B.E. degree from North China Electric Power University in 2015 and the Ph.D. degree from University of Chinese Academy of Sciences (UCAS) in 2020. She is now an Associate Professor at the Laboratory of Cognition and Decision Intelligence for Complex Systems, Institute of Automation, Chinese Academy of Sciences (CASIA). Her research interests include pattern recognition and computer vision.
\end{IEEEbiography}

\vspace{-12mm}
\begin{IEEEbiography}[{\includegraphics[width=1in,height=1.25in,clip,keepaspectratio]{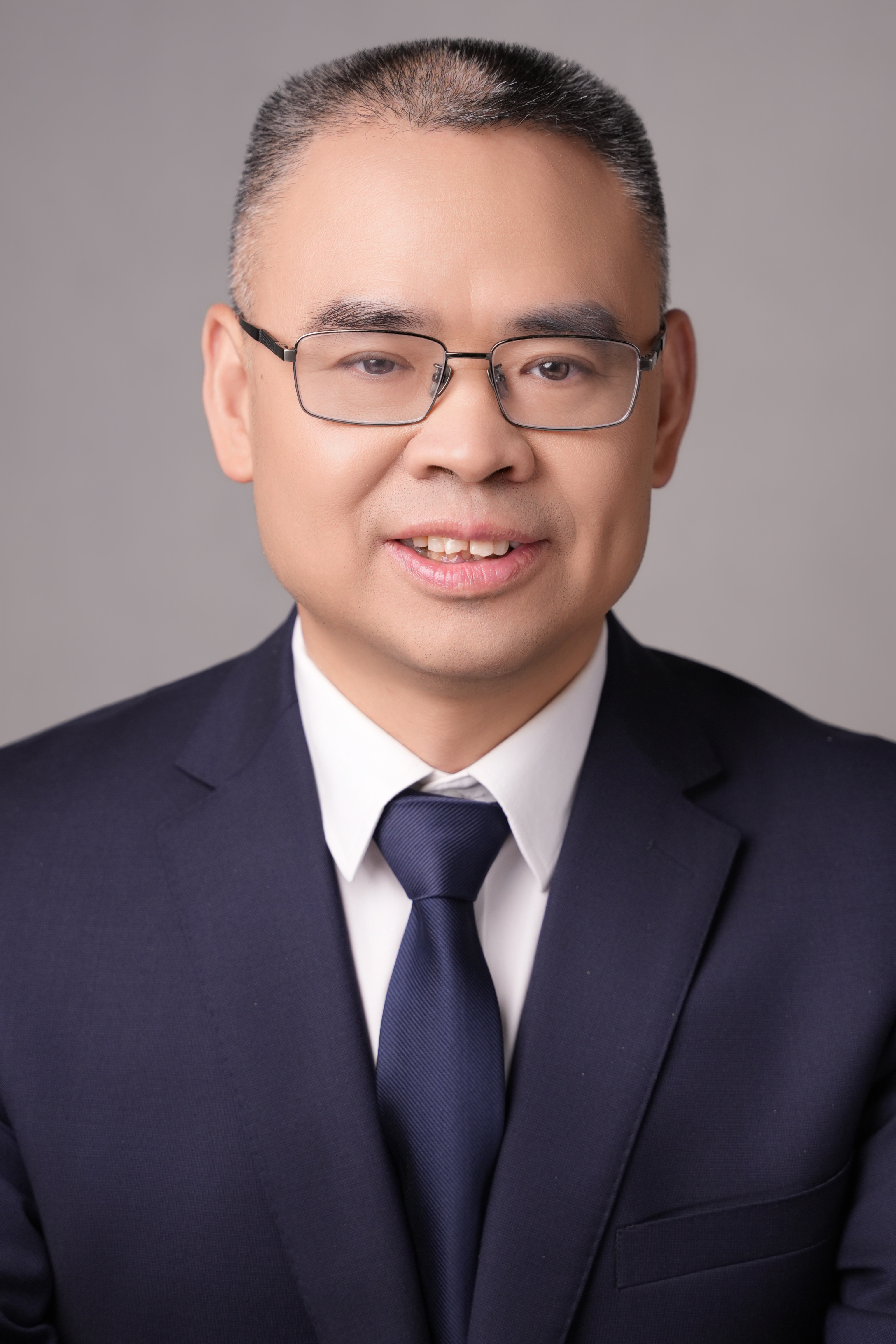}}]{Qiulin Huang}
	received the B.E. degree from Hengyang Medical College in 1996 ,and the Ph.D. degree from Central South University in 2004. He is an Professor of The First Affiliated Hospital, University of South China. His research interests include pattern recognition and computer vision.
\end{IEEEbiography}

\vspace{-12mm}
\begin{IEEEbiography}[{\includegraphics[width=1in,height=1.25in,clip,keepaspectratio]{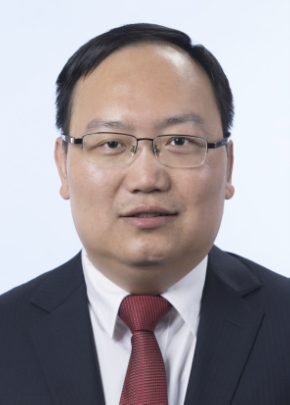}}]{Zhenan Sun}
	received the Ph.D. degree from the Institute of Automation, Chinese Academy of Sciences (CASIA) in 2006.
	He is a professor at New Laboratory of Pattern Recognition (NLPR), State Key Laboratory of Multimodal Artificial Intelligence Systems (MAIS), CASIA.
	His current research interests include biometrics, pattern recognition, and computer vision. He is a fellow of the IAPR, and an Associate Editor of the IEEE Transactions on Biometrics, Behavior, and Identity Science.
\end{IEEEbiography}

\vspace{-12mm}
\begin{IEEEbiography}
	[{\includegraphics[width=1in,height=1.25in,clip,keepaspectratio]{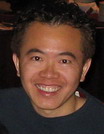}}]
	{Ming-Hsuan Yang} is a Professor of Electrical Engineering and Computer Science at University of California, Merced. Yang serves as a program co-chair of IEEE International Conference on Computer Vision (ICCV) in 2019.
	He received the Best Paper Award at ICML 2024, Longuet-Higgins Prize at CVPR 2023, NSF CAREER award in 2012
	and Google Faculty Award in 2009. He is a Fellow of the IEEE and ACM.
\end{IEEEbiography}

\end{document}